\documentclass[default,iicol]{sn-jnl}

\jyear{2021}%

\usepackage{colortbl}
\usepackage{graphicx}
\usepackage{amssymb}
\usepackage{enumerate}
\usepackage{hyperref}
\usepackage{array}
\usepackage{color}
\usepackage{multirow}
\usepackage{amsmath}
\usepackage[misc]{ifsym}
\usepackage{bbding}
\usepackage{epstopdf}

\theoremstyle{thmstyleone}%
\theoremstyle{thmstyletwo}%

\theoremstyle{thmstylethree}%

\begin{document}

\title[Article Title]{Benchmarking Deep Models for Salient Object Detection}


\author[1]{\fnm{Huajun} \sur{Zhou}}\email{zhouhj26@mail2.sysu.edu.cn}

\author[1]{\fnm{Yang} \sur{Lin}}\email{liny239@mail2.sysu.edu.cn}

\author[1]{\fnm{Lingxiao} \sur{Yang}}\email{yanglx9@mail.sysu.edu.cn}

\author[1,2]{\fnm{Jianhuang} \sur{Lai}}\email{stsljh@mail.sysu.edu.cn}

\author*[1,2]{\fnm{Xiaohua} \sur{Xie}}\email{xiexiaoh6@mail.sysu.edu.cn}

\affil[1]{\orgdiv{School of Computer Science and Engineering}, \orgname{Sun Yat-sen University}, \orgaddress{\city{Guangzhou}, \postcode{510006}, \country{China}}}
\affil[2]{\orgdiv{Key Laboratory of Machine Intelligence and Advanced Computing}, \orgname{Ministry of Education}, \orgaddress{\city{Guangzhou}, \postcode{510006}, \country{China}}}





\abstract{In recent years, deep network-based methods have continuously refreshed state-of-the-art performance on Salient Object Detection (SOD) task. However, the performance discrepancy caused by different implementation details may conceal the real progress in this task. Making an impartial comparison is required for future researches. To meet this need, we construct a general SALient Object Detection (SALOD) benchmark to conduct a comprehensive comparison among several representative SOD methods. Specifically, we re-implement 14 representative SOD methods by using consistent settings for training. Moreover, two additional protocols are set up in our benchmark to investigate the robustness of existing methods in some limited conditions. In the first protocol, we enlarge the difference between objectness distributions of train and test sets to evaluate the robustness of these SOD methods. In the second protocol, we build multiple train subsets with different scales to validate whether these methods can extract discriminative features from only a few samples. In the above experiments, we find that existing loss functions usually specialized in some metrics but reported inferior results on the others. Therefore, we propose a novel Edge-Aware (EA) loss that promotes deep networks to learn more discriminative features by integrating both pixel- and image-level supervision signals. Experiments prove that our EA loss reports more robust performances compared to existing losses. Code is available at \url{https://github.com/moothes/SALOD}.}




\maketitle

\section{Introduction}\label{sec1}

\begin{table*}
\renewcommand\arraystretch{1.1}
\setlength\tabcolsep{3pt}
\caption{The experimental settings of fourteen available models in their original paper and our re-implementations.
Back.: Backbone; V: VGG; R: ResNet; Both: VGG-16+ResNet-50; Union: a union set of MSRA-B \cite{msra} and DUT-O \cite{DUT-OMRON}; MS: multi-scale training.}
\label{tab:setting}
\centering
\footnotesize
\begin{tabular}{l|c||cccc|ccc|ccc}
\toprule
\multirow{2}{*}{Methods} & \multirow{2}{*}{Year} & \multicolumn{7}{c}{Original implementation}& \multicolumn{3}{|c}{Our implementation}   \\\cline{3-12}

&  & Back. & Train set & Train input  & Test input & MS   & Crop   & Rotate  & Back. & Train set & Train \& test inputs     \\
\midrule
DHSNet  \cite{dhsnet} & 2016  & V-16   & Union    & $224\times 224$ & $224\times 224$ &        & \checkmark &            & Both & DUTS-TR  & $320\times 320$     \\
Amulet  \cite{amulet} & 2017  & V-16   & MSRA-10K & $256\times 256$ & $256\times 256$ &        &        & \checkmark               & Both & DUTS-TR  & $320\times 320$     \\
NLDF    \cite{nldf}   & 2017  & V-16   & MSRA-B   & $352\times 352$ & $352\times 352$ &        &                   &   & Both & DUTS-TR  & $320\times 320$       \\
SRM     \cite{srm}    & 2017  & R-50   & DUTS-TR  & $353\times 353$ & $353\times 353$ &        &    &       & Both & DUTS-TR  & $320\times 320$       \\
PicaNet \cite{picanet}& 2018  & Both   & DUTS-TR  & $224\times 224$ & $224\times 224$ &        & \checkmark  &        & Both & DUTS-TR  & $320\times 320$         \\
DSS     \cite{dss}    & 2018  & Both   & DUTS-TR  & original        & $400\times 300$ &       &     &                     & Both & DUTS-TR & $320\times 320$          \\
BASNet  \cite{basnet} & 2019  & R-34   & DUTS-TR  & $224\times 224$ & $256\times 256$ &       & \checkmark  &     & Both & DUTS-TR  & $320\times 320$      \\
CPD     \cite{cpd}    & 2019  & Both   & DUTS-TR  & $352\times 352$ & $352\times 352$ &       &       &                    & Both & DUTS-TR  & $320\times 320$     \\
PoolNet \cite{poolnet}& 2019  & Both   & DUTS-TR  & original        & $400\times 300$ &       &       &                  & Both & DUTS-TR  & $320\times 320$  \\
EGNet   \cite{egnet}  & 2019  & Both   & DUTS-TR  & original        & original        &       &       &                    & Both & DUTS-TR  & $320\times 320$      \\
SCRN    \cite{scrn}   & 2019  & R-50   & DUTS-TR  & $352\times 352$ & $352\times 352$ & \checkmark       &       &                    & Both & DUTS-TR  & $320\times 320$     \\
GCPA    \cite{gcpa}   & 2020  & R-50   & DUTS-TR  & $288\times 288$ & $320\times 320$ &        & \checkmark      &                    & Both & DUTS-TR  & $320\times 320$     \\
ITSD    \cite{itsd}   & 2020  & Both   & DUTS-TR  & $288\times 288$ & $288\times 288$ & \checkmark       &       &                     & Both & DUTS-TR  & $320\times 320$      \\
MINet   \cite{minet}  & 2020  & Both   & DUTS-TR  & $320\times 320$ & $320\times 320$ &        &       & \checkmark                    & Both & DUTS-TR   & $320\times 320$      \\
\bottomrule
\end{tabular}
\end{table*}

\label{intro}
An impressive mechanism of human vision system is the internal process that quickly scans the global image to obtain region of interest.
Pioneer work \cite{rsa} was proposed to simulate this natural phenomenon.
In the field of computer vision, this task is referred to as Salient Object Detection (SOD).
Many researchers have been involved in SOD researches because (1) SOD can supply a viewpoint for better comprehending the internal process of human vision, and (2) SOD is often treated as a fast preprocessing step for assisting other computer vision techniques to achieve a more robust system.

In the last two decades, numerous efforts are devoted to making substantial advancements in SOD, \textit{e.g.}, new datasets \cite{msra} \cite{DUTS} \cite{SOC}, metrics \cite{F-score} \cite{Smeasure} \cite{Emeasure} and SOD methods \cite{AC} \cite{HC} \cite{lgpr}.
Recently, due to the robustness of Convolutional Neural Networks (CNNs) in feature extraction, SOD performance is significantly improved by CNN-based methods, including sequential networks (Long et al. \cite{fcn}; Lin et al. \cite{dcl}; Wang et al. \cite{rfcn}), Directed Acyclic Graph (DAG) networks (Zhang et al. \cite{amulet}; Zhang et al. \cite{ucf}; Liu et al. \cite{dhsnet}; Chen et al. \cite{ra}), and advanced objective functions (Feng et al. \cite{afnet}; Qin et al. \cite{basnet}; Chen et al. \cite{contour}).

Although CNN-based SOD methods have reported impressive accuracy, two challenges hinder us from revealing the real progress.
First, different SOD methods employ various implementation settings, where the discrepancy of such settings may conceal the relative performance gains.
In Tab. \ref{tab:setting}, we show the main differences between several representative methods, including backbones, train sets, input sizes, and data augmentation strategies.
Second, to the best of our knowledge, rare efforts are dedicated to investigating the robustness of existing methods in some limited conditions, such as very few training samples or large objectness shifting between train and test sets.

\textbf{To this end, we propose the SALient Object Detection (SALOD) benchmark based on a hybrid dataset to evaluate 14 representative SOD methods in terms of effectiveness, efficiency, and generalization.}
The key characteristics of the proposed benchmark are as follows:
\begin{enumerate}[1)]
\item \textbf{Dataset}.
A hybrid dataset is built based on eight prevalent SOD datasets, including SOD \cite{SOD}, PASCAL-S \cite{PASCAL-S}, ECSSD \cite{ecssd}, HKU-IS \cite{HKU-IS}, MSRA-B \cite{msra}, DUTS \cite{DUTS}, DUT-OMRON \cite{DUT-OMRON}, and THUR15k \cite{THUR15K}. Meanwhile, we use an image retrieval algorithm to remove the repeated images.
\item \textbf{Metric}.
We adopt six prevalent metrics for a comprehensive comparison, including MAE, max-$F_{\beta}$, ave-$F_{\beta}$, Fbw, E-Measure and S-Measure.
\item \textbf{Fairness}.
To conduct a robust and meaningful evaluation, we re-implement all compared methods using the same image resolution, backbone, and data augmentation strategies.
Other important factors, such as optimizer and learning rate, are tuned independently to achieve better performance.
\item \textbf{Objectness Shifting Validation}.
Objectness affects the SOD performance and follows similar distributions in existing datasets due to random splitting.
We validate the performance of existing SOD methods when a large distribution gap is established between train and test sets.
\item \textbf{Few-shot Learning}.
Deep networks benefit from rich manual annotations in experimental datasets.
However, these annotations are scarce in some industrial applications.
We construct different scales of train sets, from 10 to 10k images, to validate the robustness of several representative methods on few-shot learning.
\end{enumerate}

We draw some conclusions from our benchmark.
First, some earlier works perform significantly better than their original papers with our settings.
For example, BASNet \cite{basnet}, published in 2019 year, achieves state-of-the-art performance among available SOD methods in our benchmark with consistent settings.
Second, the performances of SOD methods are degraded if the objectness distributions are significantly different between the train and test sets.
Third, all methods have been witnessed performance drops when the number of training samples are decreased.

Another interesting finding in our benchmark is that existing loss functions are usually specialized in some metrics but reported inferior results on the others.
The main reason is that these losses fail to conclude distinctive features from different perspectives simultaneously.
For example, CTLoss \cite{contour} reports remarkable max-$F_{\beta}$ scores.
However, it is sensitive to the contours of non-salient objects and results in low ave-$F_{\beta}$ scores.
Moreover, $F_{\beta}$ loss \cite{Floss} achieves high ave-$F_{\beta}$ scores because it enlarges the distances between positive and negative features.
Since this loss is hard to distinguish the boundary pixels, it usually obtains low max-$F_{\beta}$ scores.
\textbf{To this end, we further develop a novel Edge-Aware (EA) loss to combine the advantages of these losses.
Moreover, the proposed loss promotes the networks to conclude more distinctive representations by introducing the edge information into $F_{\beta}$ term.}
Extensive experiments on our benchmark demonstrate that our EA loss substantially improves the performance of existing models on all metrics.

The remainder of this paper is structured as follows.
In Section 2, closely related works on the salient object detection task are reviewed.
In Section 3, the details of the proposed SALOD benchmark and available models, as well as experiment protocols are described.
Subsequently, we elaborate on a novel EA loss in Section 4.
Experimental results of our benchmark and the proposed EA loss are reported in Sections 5 and 6.
Finally, the paper is concluded in Section 7 .

\section{Related Works}
\subsection{Salient Object Detection Surveys}
Borji et al. \cite{Bor_survey} provided thorough descriptions about the origin and primary concepts of salient object detection task.
Furthermore, this work also reviewed the SOD history from early computational models to recent deep models.
Wang et al. \cite{wang_survey} systematically reviewed both traditional and CNN-based SOD methods.
Moreover, they conducted several experiments to validate the robustness of SOD methods, such as attribute-based evaluation, input perturbations, adversarial attack analysis, and cross-dataset validation.
Fan et al. \cite{rgbd_survey} reviewed many SOD methods on RGB-D datasets.
They proposed a new dataset to evaluate existing methods and provided some insights for future researches.
In addition, Wang et al. \cite{video_survey} summarized existing methods in the video saliency detection task.
They also investigated and analyzed various visual attention models on their benchmark.

In contrast to previous surveys, we aim at 1) evaluating many representative works under consistent settings and 2) providing some analysis in different experimental situations.
Moreover, our work reveals a limitation in current loss functions and proposes a new type of loss to improve many SOD methods.

\subsection{Salient Object Detection Datasets}
In the last several years, many datasets have been built to evaluate SOD methods.
In early datasets, researchers annotated the salient objects using bounding boxes (e.g., MSRA-B \cite{msra}).
Subsequently, they annotate images using pixel-wise binary masks (e.g., ASD \cite{THUR15K}).
Movahedi et al. attempted to reduce subjective variations by proposing the Salient Object Dataset (SOD) \cite{SOD} with 300 images and corresponding labels.
Yan et al. proposed CSSD \cite{cssd} and ECSSD \cite{ecssd} datasets to test the robustness of SOD models on small objects, especially in complex backgrounds.
Furthermore, Radhakrishna et al. \cite{THUR15K} collected images from the Internet according to several keywords to construct the THUR15K dataset.
Yang et al. \cite{DUT-OMRON} proposed the DUT-OMRON dataset that annotates multiple objects in an image, which significantly increases the difficulty of the SOD task.
Li et al. \cite{PASCAL-S} proposed a PASCAL-S dataset with non-binary masks, built on the validation set of the PASCAL VOC 2010 \cite{pascal}.
Wang et al. \cite{DUTS} provided a large-scale DUTS dataset with over 15k images in total.
Li et al. \cite{HKU-IS} constructed a more challenging HKU-IS dataset that contains low-contrast images annotated with multiple disconnected objects.
Fan et al. \cite{SOC} proposed the SOC dataset and annotated more attributes for each image, such as object size and brightness.

We collect a large-scale dataset by assembling the most prevalent SOD datasets to facilitate subsequent experiments and analyses.
Moreover, we generate various splits for different experiment protocols.
However, existing datasets encounter many issues:
1) some test sets fail to reliably evaluate SOD performance because they contains duplicate images that have been utilized to train SOD methods.
2) Non-binary masks are used in PASCAL-S dataset, but not in other ones.
To effectively evaluate SOD methods, we adopt some processes for the new datasets, such as binarizing annotations and removing repeated images.

\subsection{Evaluation Metrics}
\textbf{Mean Absolute Error (MAE)} computes the pixel-wise difference between predictions and ground truths.
Its formula can be denoted as:
\begin{equation}
    MAE = \frac{1}{W \times H}\sum_{i=1}^{W \times H} | x_{i} - y_{i}|,
\end{equation}
where $x_{i}$, $y_{i}$ indicate the predictions and labels respectively, while $W$ and $H$ are the spatial size of images.

Furthermore, $F_{\beta}$ score is employed to evaluate predicted saliency maps.
Its formula is listed as follows:
\begin{equation}
\label{eqn:fb}
F_{\beta} = \frac{(1+\beta^{2})\times p \times r}{\beta^{2}\times p + r},
\end{equation}
where $\beta^{2}$ is set to 0.3 in general \cite{THUR15K}.
The \textbf{maximum (max) $F_{\beta}$} and \textbf{average (ave) $F_{\beta}$} are computed by Eqn. \ref{eqn:fb} with different positive thresholds from 0 to 255.
$p$ and $r$ denote precision and recall, respectively, which can be computed by:
\begin{equation}
    p = \frac{TP}{TP + FP}, \qquad
    r = \frac{TP}{TP + FN},
\end{equation}
where TP, FP, and FN are true-positive, false-positive, and false-negative, respectively.

Margolin et al. \cite{fbw} assigned different weights to different misclassified pixels according to different positions and contexts.
Their proposed metric is called \textbf{weighted $F_{\beta}$ measure} (Fbw) \cite{fbw}, which is formulated as follows:
\begin{equation}
F_{\beta}^{w} = \frac{(1+\beta^{2})\times p^{w} \times r^{w}}{\beta^{2}\times p^{w} + r^{w}},
\end{equation}
where $w$ indicates different weights according to spatial positions of pixels.

Pan et al. \cite{Smeasure} proposed a \textbf{Similarity Measure (S-Measure)} that considers both region-aware and object-aware structural similarities as follows:
\begin{equation}
SM = \alpha \times S_{o} + (1 - \alpha) \times S_{r},
\end{equation}
where $\alpha$ set to 0.5.
$S_{o}$ and $S_{r}$ are object-aware and region-aware structural similarities, respectively.

Pan et al. \cite{Emeasure} further proposed \textbf{Enhanced-alignment Measure (E-Measure)} to capture global statistics and the local pixel matching information:
\begin{equation}
EM = \frac{1}{W \times H} \sum_{i=1}^{W}\sum_{j=1}^{H} \phi (i, j),
\end{equation}
where $\phi(i, j)$ is the enhanced alignment matrix.

Those metrics are prevalent in SOD evaluations.
However, most existing works do not report scores on all of them.
Based on our experiment, these methods specialize in one or several metrics but are inferior on the other metrics.
Therefore, we include all these metrics in our benchmark to provide a detailed comparison.

\subsection{Salient Object Detection Methods}
Some early works \cite{LC}\cite{HC} followed the intuition that salient pixels are more distinctive with the global context in color space.
Achanta et al. \cite{AC} argued that salient regions are related to their contrast with their neighborhoods at various scales.
Li et al. \cite{dsr} took advantage of the prior knowledge that most image boundary superpixels are background.
Thus, they reconstructed the saliency maps by matching inner superpixels with boundary ones.
Jiang et al. \cite{mc} used these boundary superpixels as the initial nodes in a Markov chain.
Then, internal nodes merge with the most similar node and update the new node.
Jiang et al. \cite{ufo} proposed a method that integrates uniqueness, focusness, and objectness to detect salient regions.
Duan et al. \cite{swd} measured the saliency based on three related factors: the dissimilarities between image patches, the spatial distances between image patches, and the central bias.
Jiang et al. \cite{cb} analyzed pixels' context based on multi-scale superpixels to compute coarse saliency maps.
Subsequently, they extract object-level shape priors to combine saliency with object boundary information.
Finally, the saliency predictions are generated by iteratively updating the above coarse saliency maps and object-level shape priors.

Since CNN-based methods have surpassed hand-crafted methods recently, we will present CNN-based SOD methods in this paper and refer interested readers to surveys  \cite{Bor_survey} \cite{sod_survey} \cite{wang_survey} for more traditional methods.
Existing CNN-based methods can be divided into two groups according to the network structure, including Sequential networks and Directed Acyclic Graph (DAG) networks.
We review some representative methods of these two groups in the following paragraphs.

\paragraph{\textbf{Sequential Networks.}}
We define that a sequential network only contains a feed-forward structure without skip-connections for simplicity.
Some researchers utilize such kinds of networks as generic feature extractors to conclude distinctive representations.

He et al. \cite{supercnn} combined predictions from multiple scales of superpixels to form a smooth saliency map.
For each scale, they extracted a color uniqueness sequence and a color distribution sequence from each superpixel.
Several CNNs are employed to generate the final predictions based on these sequences.
Wang et al. \cite{legs} used two subnetworks to generate local estimation and object proposals, respectively.
After that, they utilized a saliency prediction network to extract a feature from each proposal and output a corresponding confidence score.
Finally, these proposals are fused with the local estimation to generate the predictions.
Zhao et al. \cite{gl} utilized CNNs to model both global and local contexts using two different scales windows.
Lee et al. \cite{super} predicted an encoded low-level distance map for each superpixel using a deep network.
In addition, they extracted a global feature using the VGG-16 \cite{vgg} network for each image.
Then, all distance maps are integrated with the global features to discriminate whether this superpixel is salient.
Zhang et al. \cite{imlc} employed a deep network to output a coarse saliency map for each image.
They refined this low-resolution saliency map by using three scales of over-segmented images produced by the SLIC algorithm \cite{slic}.

\paragraph{\textbf{DAG Networks.}}
Deep networks have been witnessed remarkable advancements owing to the introduction of skip connection, forming the Directed Acyclic Graph (DAG) network.
Long et al. \cite{fcn} transformed the fully-connected layers in VGG-16 into convolutional layers to predict a coarse saliency map.
Then, they upsample the feature maps from the topmost layers to generate a coarse-to-fine prediction.
Li et al. \cite{ilsod} generated multiple salient object proposals based on the predicted contour maps to implement an instance-level salient object segmentation.
Li et al. \cite{dcl} introduced a network that consists of a pixel-level fully convolutional stream and a segment-wise spatial pooling stream.
Tang et al. \cite{dscn} integrated five features from different stages of the Deeply Supervised Net (DSN) \cite{dsn} to predict the final saliency maps.
Wang et al. \cite{srm} outputted coarse predictions using a transformed ResNet-50.
Then, another truncated ResNet-50 is employed to produce larger feature maps to refine the above coarse predictions.
Wu et al. \cite{cpd} cascaded two partial decoders to refine the predictions sequentially.
These partial decoders only utilized multiple high-level encoder features and thus substantially reduce the computation loads.

Except for developing various network structures, most recent works focus on enhancing the U-shape network \cite{unet}, which has reported impressive results on multiple segmentation tasks.
In this network, the learned feature maps are upsampled to input size progressively.
Skip connections are attached to the network to integrate both low-level and high-level representations.
Liu et al. \cite{dhsnet} upgraded the U-shape structure by using the Recurrent Convolutional Layer (RCL) \cite{rcl} as its decoder blocks to enlarge the feature maps step by step.
Zhang et al. \cite{amulet} improved the U-shape structure by fusing multiple intermediate maps to generate refined predictions.
Inspired by \cite{nest}, Hou et al. \cite{dss} employed a series of skip connections to build a strong relationship between each pair of side paths.
Liu et al. \cite{picanet} scanned over images from four directions to conclude global and local cues using multiple LSTMs \cite{lstm}.
Zhang et al. \cite{pagrn} utilized spatial-wise and channel-wise attention modules to enhance the learned features sequentially.
Zhang et al. \cite{bmp} changed the skip connection to a bi-directional feature passing module.
Zeng et al. \cite{inv} extended the receptive fields of convolutional blocks to strengthen the learned representations by integrating multi-scale features.
Wang et al. \cite{page} introduced pyramid attention modules and salient edge detection modules to improve the skip-connections of the U-shape structure.
Chen et al. \cite{ra} proposed reverse attention that erases the current prediction to discover the missing object parts.
Zhao et al. \cite{pfa} developed a two-pathway network to learn low-level and high-level features simultaneously.
Wang et al. \cite{PFPN} proposed a feature polishing module to progressively polish the multi-level features to be more accurate and representative.
Chen et al. \cite{gcpa} utilized a global context-aware progressive aggregation network to integrate both low-level, high-level, and global information.
Wei et al. \cite{f3net} synthesized the local structure information of pixels to guide the network to focus more on local details.
Pang et al. \cite{minet} introduced the aggregate interaction module to better fuse the learned features in the decoder.

Instead of only use saliency as supervision signals, some works attempt to introduce contour information to assist the SOD models.
Li et al. \cite{ckt} proposed an iterative structure that utilizes contour and saliency cues to guide the learning process of each other alternately.
Feng et al. \cite{afnet} utilized contour cues to supervise the edges of saliency predictions.
Wu et al. \cite{mlm} proposed to train deep networks for saliency detection using multi-task intertwined supervision.
Under this supervision, they employ foreground contour detection to improve the accuracy of the saliency detection task.
Wu et al. \cite{scrn} developed a network that bidirectionally passes messages between salient object detection and edge detection.
Liu et al. \cite{poolnet} utilized edge information as supervision signals for the decoder features in different stages.
Zhao et al. \cite{egnet} forced the topmost feature in the U-shape structure to learn the edge maps of input images.
After that, this feature is integrated with additional features from other stages to generate saliency predictions.

\paragraph{\textbf{Loss Function.}}
Besides the network design, developing new objective functions also attracts increasing attention.
In most SOD models, they apply Binary Cross-Entropy (BCE) loss to train the networks.
It can be calculated by:
\begin{equation}
L_{BCE} = - \frac{1}{W \times H}\sum_{i=1}^{W \times H} (y_ilogx_i + (1-y_i)lox(1-x_i)),
\end{equation}
where $x_{i}$ and $y_{i}$ are the prediction and ground truth, respectively.
Chen et al. \cite{contour} argued that pixels closed to boundaries are hard examples, so they applied higher weights to these pixels during the training process.
This loss can be computed by:
\begin{align}
L_{c} &= - \frac{1}{W \times H}\sum_{i=1}^{W \times H} \gamma_{i}(y_ilogx_i + (1-y_i)lox(1-x_i)),\\
\gamma_{i} &= (max(x^{c}_i, y^{c}_i) \times k + 1),
\end{align}
where $k$ is a hyperparameter and $\gamma_{i}$ is the weight for the $i$-th pixel.
$x^{c}_i$ and $ y^{c}_i$ indicate the predicted and annotated contour masks, respectively.

Luo et al. \cite{nldf} borrowed the Dice loss from \cite{dice} to impose statistical supervision on the final prediction.
Its formula is:
\begin{equation}
L_{Dice} = 1 - \frac{2 TP}{(FN + TP) + (FP + TP)}.
\end{equation}
Qin et al. \cite{basnet} appended the Structural SIMilarity (SSIM) and Intersection-over-Union (IOU) losses to the traditional BCE loss, where SSIM loss and IOU loss are shown as follows:
\begin{align}
L_{SSIM} &= 1 - \frac{(2\mu_{x}\mu_{y} + C_{1})(2\sigma_{xy} + C_{2})}
{(\mu_{x}^{2}\mu_{y}^{2} + C_{1})(\sigma_{x}^{2} + \sigma_{y}^{2} + C_{2})},\\
L_{IOU} &= 1 - \frac{TP}{FN + TP + FP},
\end{align}
where $\mu_{x}$, $\mu_{x}$ and $\sigma_{x}$, $\sigma_{y}$ are the mean and standard deviations of $x$ and $y$ respectively, $\sigma_{xy}$ is their covariance.
C1 = 0.012 and C2 = 0.032 are used to avoid dividing by zero.
Zhao et al. \cite{Floss} suggested to directly use the evaluation metric $F_{\beta}$ as the objective function to train the network.
Its formula is shown as follows:
\begin{equation}
\label{equ:lf}
L_{f} = 1 - \frac{(1+\beta^{2}) TP}{(FN + TP) + \beta^{2}(FP + TP)}.
\end{equation}

These objective functions provide various supervision signals to networks and have shown impressive performance in previous works.
However, existing loss functions are insufficient to guide deep networks to achieve significant performance on all metrics.
Therefore, we develop a novel loss to combine the advantages of existing loss functions.

\begin{figure*}[t]
\includegraphics[width=6.3in,height=4.0in]{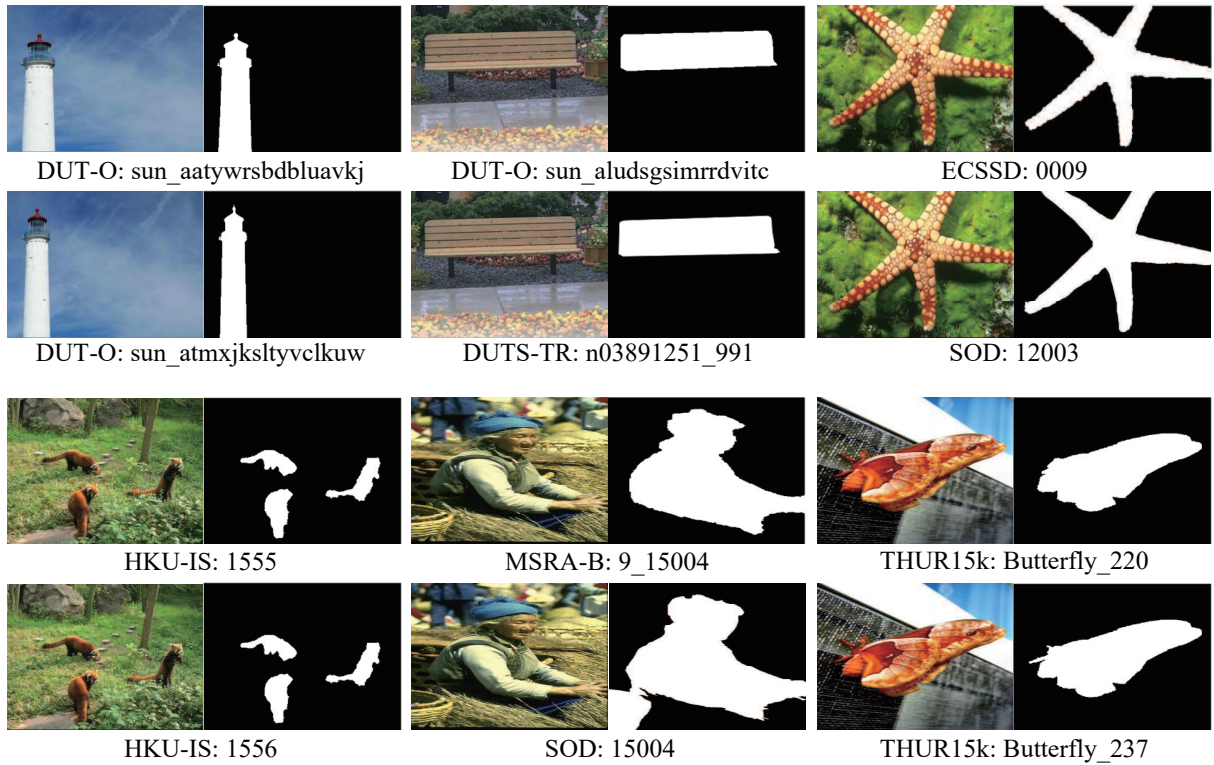}
\caption{Duplicate images in existing datasets. For each image, its original dataset and image name are marked under image.}
\label{fig:repeat}
\end{figure*}

\section{Benchmark}
In section \ref{sec:setup}, we first introduce a novel SALient Object Detection (SALOD) benchmark, which addresses some issues in existing SOD datasets.
In section \ref{sec:setting}, we elaborate on our settings to ensure an impartial comparison between SOD methods.
Finally, we present the details of three protocols in section \ref{sec:expsetup}.

\subsection{Hybrid Dataset}
\label{sec:setup}
Current datasets exist two issues that need to be addressed.
1) These datasets contains duplicate images with different annotations.
This will cause redundant evaluations and thus potentially affect many of previous works.
2) Most datasets use binary masks to annotate the salient objects in images, while PASCAL-S uses non-binary labels.
To tackle these issues and facilitate further validation experiments, we collect a large-scale hybrid SALOD dataset by gathering 38570 images from eight popular SOD datasets, including SOD \cite{SOD}, PASCAL-S \cite{PASCAL-S}, ECSSD \cite{ecssd}, HKU-IS \cite{HKU-IS}, MSRA-B \cite{msra}, DUT-OMRON \cite{DUT-OMRON}, THUR15k \cite{THUR15K}, and DUTS \cite{DUTS}.
These datasets (from the first to the last) have 300, 850, 1000, 4447, 5000, 5168, 6233, and 15572 pixel-wise labeled images, respectively.
We list more details about these datasets in Tab. \ref{tab:dataset}.
To remove duplicate images, an image retrieval method - Hnswlib \footnote{https://github.com/Sg4Dylan/EfficientIR} \cite{Hnswlib}, is employed to find the top-5 most similar images for each image.
This process produces over 192,850 image pairs in total.
Since duplicate images are very similar, we use a similarity threshold of 0.97 to find these image pairs, resulting in a total of 4396 image pairs.
We find that some images may be similar in the feature space but different in the original image space.
Thus, three subjects independently check the remainders to distinguish whether they are ``same'' or not.
Finally, we remove 640 images labeled as ``same'' by at least two subjects, resulting in a dataset with 37930 images in total.
In Fig. \ref{fig:repeat}, we show some duplicate images as well as their corresponding information.

\begin{table}[!t]
\renewcommand\arraystretch{1.1}
\setlength\tabcolsep{3pt}
\caption{Available datasets to construct the proposed SALOD dataset. }
\footnotesize
\label{tab:dataset}
\begin{tabular}{l|c|c|c|c}
\hline
Methods        & Publish.       & \#Img.  & \#Obj. & Size  \\
\noalign{\smallskip}\hline\noalign{\smallskip}
MSRA-B \cite{msra}         & CVPR2007  & 5000  & 1-2   & 126$\sim$400  \\
SOD \cite{SOD}             & CVPRW2010 & 300   & 1-4+  & 321$\sim$481    \\
DUT-OMRON \cite{DUT-OMRON} & CVPR2013  & 5168  & 1-4+  & 139$\sim$401  \\
PASCAL-S \cite{PASCAL-S}   & CVPR2014  & 850   & 1-4+  & 139$\sim$500   \\
ECSSD \cite{ecssd}         & TPAMI2015 & 1000  & 1-4+  & 139$\sim$400   \\
THUR15k \cite{THUR15K}     & TPAMI2015 & 6233  & 1-2   & 144$\sim$400  \\
HKU-IS \cite{HKU-IS}       & CVPR2015  & 4447  & 1-4+  & 100$\sim$500   \\
DUTS \cite{DUTS}           & CVPR2017  & 15572  & 1-4+  & 100$\sim$500  \\

\noalign{\smallskip}\hline
\end{tabular}
\end{table}

Based on our hybrid dataset, we introduce three protocols in our benchmark, including standard benchmark, objectness-shifting validation, and few-shot learning.
We provide various splits for different protocols.
1) In the standard benchmark experiment, we randomly select 10000 images to train available SOD methods, while the rest 27930 images are the test set.
2) Objectness distributions may affect the performance of SOD methods, so we validate SOD methods on the test sets that have different objectness distributions with train set.
Specifically, we compute the objectness score for each image and sort all images according to their scores.
The highest 10000 images are the train set, while the lowest 10000 and the rest images are two test sets with different objectness distributions.
3) In the few-shot learning experiment, we build multiple train sets with different scales.
Meanwhile, the test set of the first protocol is adopted to validate the generalization ability of SOD methods in this protocol.
For more details about these protocols please refer to section \ref{sec:expsetup}.

\subsection{Training Settings}
\label{sec:setting}
In the proposed benchmark, we adopt the following settings for all methods:
\begin{enumerate}[1)]
\item The input size is $320 \times 320$ for single-scale training.
In multi-scale training, the input size is selected from ($256 \times 256$, $320 \times 320$, $384 \times 384$) randomly;
\item We use different splits of the train and test sets for various experimental protocols.
We describe more details in section \ref{sec:expsetup};
\item In our benchmark, six frequently used metrics, including MAE, max-$F_{\beta}$, ave-$F_{\beta}$, Fbw, S-Measure, and E-Measure, are employed to evaluate the models;
\item For data augmentation, we employ horizontal flipping for all experiments.
In addition, in the first protocol, multi-scale training is adopted to prove that our re-implementations can achieve similar results as the original implementations;
\item We use ResNet-50 and VGG-16 as backbones because of their impressive generalization ability.
The ImageNet pre-trained weights are employed to initialize all backbones, the same as previous methods.
\end{enumerate}

\begin{figure*}[!t]
\includegraphics[width=6.3in,height=2.7in]{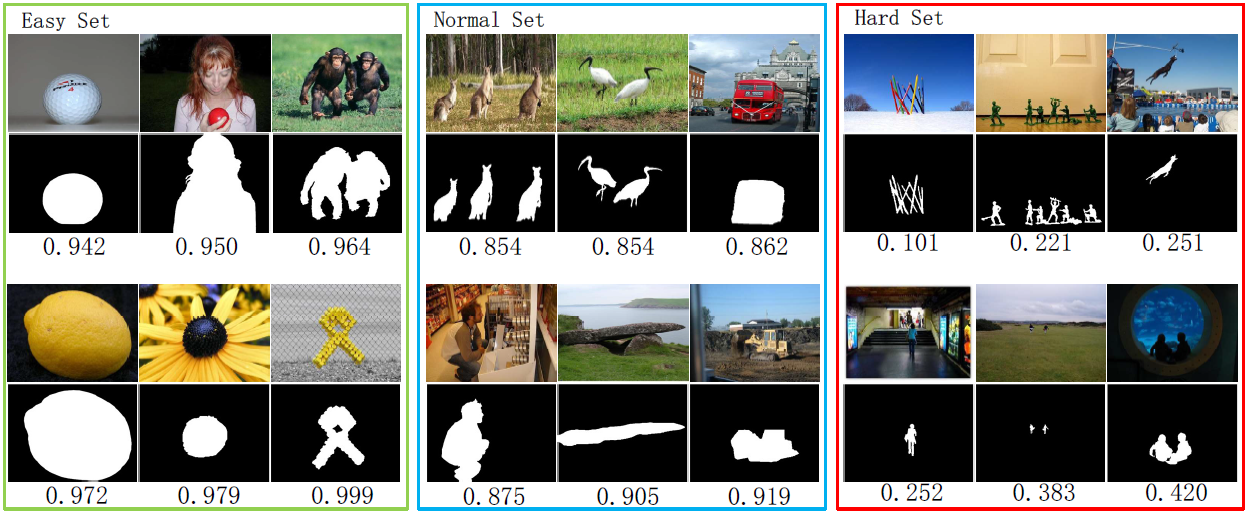}
\caption{Sample images in the proposed SALOD dataset.
Green, blue and red boxes contain images from easy, normal and hard sets, respectively.
For each sample, we show image, ground truth and objectness score from top to bottom.}
\label{fig:results}
\end{figure*}

We re-implement 14 SOD methods in our benchmark.
They are DHSNet \cite{dhsnet}, Amulet \cite{amulet}, NLDF \cite{nldf}, SRM \cite{srm}, PicaNet \cite{picanet}, DSS \cite{dss}, BASNet \cite{basnet}, CPD \cite{cpd}, PoolNet \cite{poolnet}, EGNet \cite{egnet}, SCRN \cite{scrn}, GCPA \cite{gcpa}, ITSD \cite{itsd}, MINet \cite{minet}.
Some early methods have some issues when using our settings, so we slightly modify their network structures in our re-implementations:
\begin{enumerate}[1)]
\item DHSNet \cite{dhsnet} requires a fixed input size because it flattens the topmost encoder feature and reshapes this feature to a fixed-size feature map after a fully connected layer.
This operation is not compatible with multi-scale training, so we directly reduce the channels of the topmost encoder feature to 1 using a convolutional layer.
In this way, we can train DHSNet with multi-scale inputs.
\item PicaNet \cite{picanet} uses a novel global attention module that reshapes the channel dimension to spatial size.
Since the channel dimension is fixed, this method also requires a fixed input size.
Thus, we randomly crop input images and resize them to a fixed size in multi-scale training.
\item Amulet \cite{amulet} outputs foreground and background maps simultaneously.
For simplicity, we only output a saliency map as other methods.
\item DHSNet \footnote{www.github.com/xsxszab/DHSNet-Pytorch} and SRM \footnote{www.github.com/xsxszab/SRM-Pytorch} are based on third-party implementations because no official code is available.
\end{enumerate}

\subsection{Protocol Setup}
\label{sec:expsetup}
We set up three protocols to evaluate the effectiveness of available methods under consistent settings:
1) In the standard experiment, we train these methods based on the training settings in section \ref{sec:setting} to conduct a comprehensive comparison.
2) Instead of randomly splitting the dataset, objectness-shifting validation is conducted using a new split to evaluate the generalization ability of SOD methods.
This split enlarges the difference between the objectness distributions of the train and test sets.
3) We adopt few-shot learning by selecting different scales of the train sets, from 10 to 10k images, to train available models.
In the next, we elaborate more details of these protocols.

\paragraph{\textbf{Standard Benchmark}}
\label{sec:cornerstone}
The purpose of our benchmark is to compare the effectiveness of existing SOD methods.
Therefore, we train all available methods using the settings mentioned in section \ref{sec:setting}.

We conduct a preliminary experiment to ensure that our re-implementations are comparable with the original ones.
In detail, we train these methods on the DUTS-TR dataset as the same setup in most recent works.
Using ResNet-50 as the backbone, these networks are trained with single-scale and multi-scale inputs, respectively.
To fully explore the potential of SOD methods, we tune some training factors separately, such as optimizer and learning rate schedule.
We recompute the max-$F_\beta$ and MAE scores on six prevalent test sets, including SOD, PASCAL-S, ECSSD, DUTS-TE, HKU-IS, and DUT-OMRON.

Based on the preliminary experiment, we deploy all methods on the proposed SALOD dataset.
We train all compared methods based on the pretrained VGG-16 and ResNet-50 in single-scale setting.
The optimizer and learning rate schedule for each method keeps the same as the preliminary experiment.
During testing, we adopt six metrics to evaluate the predictions of these methods, including max-$F_{\beta}$, ave-$F_{\beta}$, Fwb, MAE, S-Measure, and E-Measure.
Besides the detection performance, we also include some other criterions to measure the efficiency of these methods.
These criterions are the number of parameters (\# Param.), Multiply and ACcumulate operations (MACs), and Frame Per Second (FPS) during testing.

\begin{figure*}[]
\includegraphics[width=6.3in,height=2.9in]{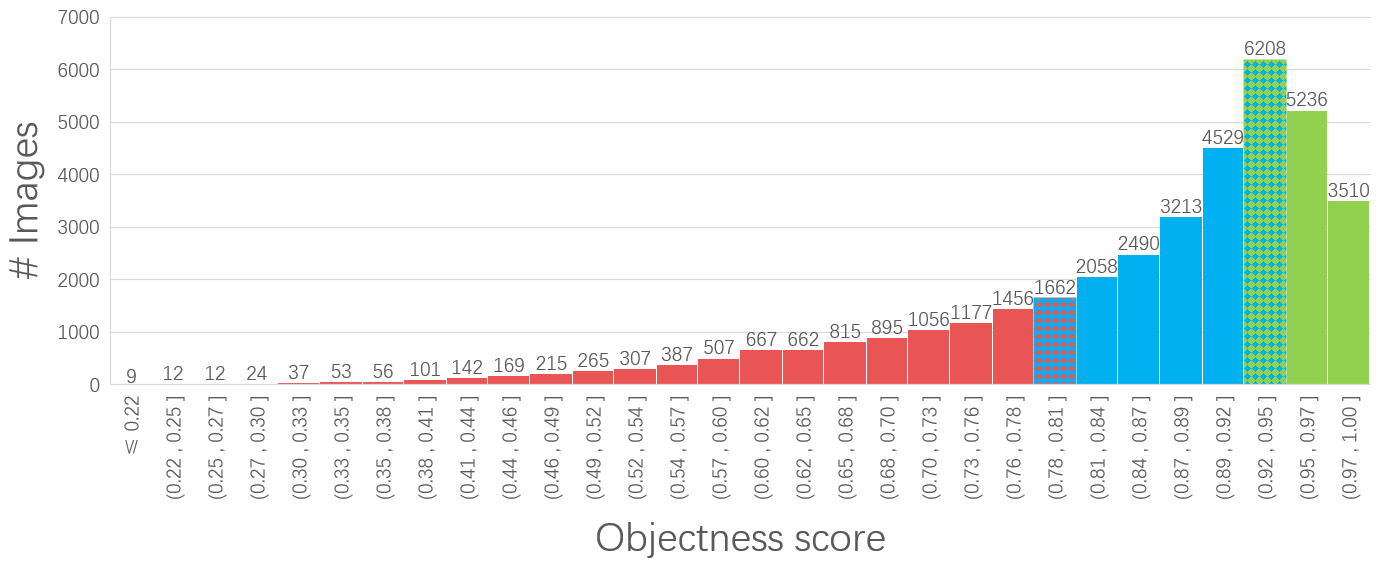}
\caption{Histogram of objectness scores.
Red, blue and green blocks indicate hard, normal and easy sets.
The x- and y-coordinates indicate the range of objectness scores and corresponding image number, respectively.}
\label{fig:score}
\end{figure*}

\paragraph{\textbf{Objectness Shifting Validation}}
Although the community has witnessed remarkable advancements in recent SOD methods, it is still unclear whether they perform well in other challenging scenarios.
One fundamental prerequisite of current SOD methods is that the training and testing data follow similar distributions.
However, this ideal hypothesis is fragile in real cases since we can hardly impose constraints on the testing data distribution.
This means that although the SOD methods minimize the empirical error on the training data, they may perform poorly on the testing data.

Some works \cite{SOC} \cite{sod_survey} concluded several challenging attributes of images that may influence the performance of SOD methods, such as heterogeneous objects, complex scenes, and small objects.
For each image, they manually annotate all attributes to facilitate the analysis of distribution shifting problem.
However, hand-crafted labels are not suitable for our benchmark because:
1) Some attributes are subjective concepts and difficult to be described using binary labels, such as complex scenes.
2) Some attributes, such as heterogeneous objects and occlusion, may not significantly degrade the performance of CNN-based SOD methods.
3) The difficulties of images are affected by multiple attributes.
Therefore, a single binary attribute can not well measure the difficulty of images.
4) Previous works only include thousands of images in their test sets.
Annotating 37930 images in our benchmark is laborious.

To this end, we propose an automatic labeling method using objectness score, which is closely related to saliency, as discussed in many previous works \cite{objectness} \cite{ufo} \cite{fgo} \cite{dhsnet}.
Windows with higher objectness scores indicate that the target objects are much more salient.
Meanwhile, large and salient objects can significantly increase the objectness scores of images.
Therefore, we use the objectness scores as measurements.

To generate objectness scores, we first use an object recognition network -- Inception-v4 \footnote{https://github.com/Cadene/pretrained-models.pytorch} \cite{inceptionv4} to generate class probability vectors.
This network is pre-trained on ImageNet \cite{imagenet} and generates 1001 probabilities for each image, including 1000 semantic classes and one background class.
We regard that class probabilities larger than 0.01 indicate these semantic classes are visible in images.
Since images may contain multiple salient objects belong to different classes, we sum the probabilities of visible classes as the objectness scores of images.
As we can see in Fig. \ref{fig:results}, images in the green box obtain high objectness scores because the main objects are very conspicuous.
Images in the red box, small objects or complex scenes result in low objectness scores.
After calculating the objectness scores, all images are sorted according to their objectness scores.
Top 10,000 images with the highest and the lowest scores are our easy and hard subsets, respectively.
The rest 17,930 images are our normal set.
Objectness distributions of three subsets are shown in Fig. \ref{fig:score}.

In our experiment, all methods are trained on the easy set and evaluated on the other two subsets.
As we can see in Fig. \ref{fig:score}, the train and test sets follow different objectness distributions instead of similar objectness distributions by random splitting.
Meanwhile, the hard set contains much more severe objectness shifting than the normal set.
Through this experiment, we examine the generalization ability of existing SOD methods and provide a new perspective to validate the effectiveness of SOD methods.

\paragraph{\textbf{Few-shot Learning}}
CNN-based SOD methods have achieved impressive performance by extracting more general representations from numerous training samples.
However, large-scale pixel-wise labeling for SOD task is laborious.
For example, we only have tens or hundreds of fully labeled images in some practical applications.
In this case, CNN models tend to overfit the training samples and produce poor predictions for unseen images.
Extracting more general features from a small number of samples is essential.
Whether these methods are sensitive to the scales of the train set is still a question.

In this section, we evaluate many methods with a limited number of training samples.
For data splits, ten scales of train sets are constructed in this protocol, including 10, 30, 50, 100, 300, 500, 1k, 3k, 5k, and 10k images, respectively.
For SOD methods, we choose some representative methods in our experiment, including DHSNet \cite{dhsnet}, Amulet \cite{amulet}, SRM \cite{srm}, DSS \cite{dss}, BASNet \cite{basnet}, CPD \cite{cpd}, PoolNet \cite{poolnet}, ITSD \cite{itsd} and MINet \cite{minet}.
We use ten train sets to train these methods and report results on six metrics (max-$F_\beta$, ave-$F_\beta$, Fbw, MAE, S-Measure, and E-Measure).

\begin{table*}
\definecolor{mycyan}{cmyk}{.3,0,0,0}
\renewcommand\arraystretch{1.1}
\setlength\tabcolsep{3pt}
\caption{Results of preliminary experiment. For each method, the first line shows the results from official saliency maps. The second and third lines list results of our implementation with single-scale (ss) and multi-scale (ms) training respectively. 
}
\label{tab:reproduce}
\centering
\footnotesize
\begin{tabular}{c|cc|cc|cc|cc|cc|cc|cc}
\noalign{\smallskip}\toprule
\multirow{2}{*}{Methods}  & \multirow{2}{*}{Back.}  & \multirow{2}{*}{Src} & \multicolumn{2}{c|}{SOD}   & \multicolumn{2}{c|}{PASCAL}   & \multicolumn{2}{c|}{ECSSD} & \multicolumn{2}{c|}{HKU-IS} & \multicolumn{2}{c|}{DUTS-TE} & \multicolumn{2}{c}{DUT-O}     \\\cline{4-15}
&&& max-$F_\beta$ & MAE & max-$F_\beta$ & MAE & max-$F_\beta$ & MAE & max-$F_\beta$ & MAE & max-$F_\beta$ & MAE & max-$F_\beta$ & MAE\\
\noalign{\smallskip}\midrule
         & V-16     & orig. & .827 & .128 & .820 & .091 & .906 & .059 & .890 & .053  & .808 &.067 & --  & --  \\
DHSNet \cite{dhsnet} & R-50  & ss    & .868 & .102 & .865 & .070    & .940 & .041 & .930 & .034  & .870 & .044    & .796   & .061   \\
         & R-50  & ms    & .873 & .104 & .864 & .070    & .943 & .041 & .930 & .035  & .880 & .044   & .807     & .059 \\\hline
\rowcolor{mycyan}  & V-16     & orig. & .798 & .144 & .828 & .100 & .915 & .059 & .897 & .051  & .778 & .085   & .743   & .098   \\
\rowcolor{mycyan} Amulet \cite{amulet} & R-50  & ss    & .868 & .107  & .856 & .070    & .933 & .044 & .924 & .036  & .860    & .045 & .781    & .063  \\
\rowcolor{mycyan}         & R-50  & ms    & .861 & .112  & .861 & .070    & .933 & .046  & .925 & .038  & .867 & .045   & .788     & .061 \\\hline
         & V-16     & orig. & .841 & .125 & .822 & .098     & .905 & .063  & .902 & .048  & .813 & .065    & .753    & .080  \\
NLDF   \cite{nldf}   & R-50  & ss    & .866 & .103 & .861 & .068    & .933 & .041 & .923 & .034  & .867 & .042   & .792  & .058    \\
         & R-50  & ms    & .865 & .106 & .860 & .069    & .936 & .042 & .925 & .034  & .875 & .041   & .804   & .057   \\\hline
\rowcolor{mycyan}         & R-50  & orig. & .843 & .128 & .838 & .084    & .917 & .054  & .906 & .046  & .826 & .059   & .769   & .069   \\
\rowcolor{mycyan} SRM    \cite{srm}    & R-50  & ss    & .841 & .119 & .835 & .080    & .919 & .052 & .901 & .046  & .824 & .055   & .763   &  .068 \\
\rowcolor{mycyan}         & R-50  & ms    & .847 & .114 & .840 & .083    & .922 & .053 & .904 & .048  & .832 & .058   & .773     & .071 \\\hline
         & R-50  & orig. & .856 & .104 & .857 & .076    & .935 & .046  & .918 &.043  & .860 & .051   & .803 & .065   \\
PicaNet\cite{picanet}& R-50  & ss    & .867 & .104 & .857 & .074    & .934 & .044 & .923 & .038  & .869 & .045   & .797  & .063    \\
         & R-50  & ms    & .860 & .105 & .857 & .075    & .933 & .049 & .917 & .043  & .867 &.048    & .800    & .065  \\\hline
\rowcolor{mycyan}         & R-50  & orig. & .846 & .124 & .831 & .093    & .921 & .052  & .900 & .050  & .826 &.065    & .769 & .063   \\
\rowcolor{mycyan} DSS    \cite{dss}    & R-50  & ss    & .846 & .113 & .852 & .073    & .930 & .048 & .913 & .041  & .848 & .048   & .776   & .060   \\
\rowcolor{mycyan} & R-50  & ms    & .849 & .112 & .855 & .075    & .928 & .051 & .913 & .044  & .859 & .050   & .793    & .065 \\\hline
         & R-34  & orig. & .851 & .114 & .854 & .076    & .942 & .037 & .928 & .032  & .859 & .048   & .805     & .056 \\
BASNet\cite{basnet}  & R-50  & ss    & .874 & .095 & .866 & .062    & .949 & .033 & .937 & .027  & .889 & .036   & .818   & .051  \\
         & R-50  & ms    & .884 & .089 & .869 & .060    & .951 & .032 & .938 & .028  & .894 & .034   & .821    & .051  \\\hline
\rowcolor{mycyan}         & R-50  & orig. & .860 & .112 & .859 & .071    & .939 & .037 & .925 & .034  & .865 & .043   & .797   & .056 \\
\rowcolor{mycyan} CPD   \cite{cpd}     & R-50  & ss    & .860 & .110  & .867 & .068    & .937 & .043 & .925 & .036  & .871    & .043 & .798    & .059  \\
\rowcolor{mycyan}         & R-50  & ms    & .866 & .109 & .871 & .069    & .941 & .042 & .928 & .038  & .876 & .044   & .809     & .059 \\\hline
         & R-50  & orig. & .871 & .102  & .863 & .075    & .944 & .039 & .931 & .034  & .880 & .040    & .808 & .056   \\
PoolNet\cite{poolnet}& R-50  & ss    & .870 & .103 & .870 & .063    & .940 & .040  & .931 & .033  & .881    & .039 & .798    & .057  \\
         & R-50  & ms    & .870 & .098 & .871 & .064    & .943 & .038 & .933 & .033  & .883 & .040   & .812     & .056 \\\hline
\rowcolor{mycyan}         & R-50  & orig. & .880 & .099 & .865 & .074    & .947 & .037 & .934 & .032  & .889 & .039   & .815     & .053 \\
\rowcolor{mycyan} EGNet  \cite{egnet}  & R-50  & ss    & .870 & .104 & .863 & .069    & .946 & .039 & .930 & .034  & .879 & .043   & .811    & .058  \\
\rowcolor{mycyan}         & R-50  & ms    & .871 & .104 & .866 & .068    & .949 & .038 & .930 & .035  & .887  &.042   & .815    & .057  \\\hline
         & R-50  & orig. & .867 & .107 & .877 & .063    & .950 & .037  & .934 & .034  & .888 & .040    & .811    & .056  \\
SCRN  \cite{scrn}    & R-50  & ss    & .878 & .096 & .869 & .065    & .943 & .039 & .932 & .034  & .881 & .042   & .807    & .060  \\
         & R-50  & ms    & .875 & .100 & .873 & .065    & .944 & .040 & .932 & .036  & .887 & .041   & .812    & .059  \\\hline
\rowcolor{mycyan}         & R-50  & orig. & .876 & .090 & .869 & .062    & .948 & .035  & .938 & .031  & .888 &.038    & .812 & .056  \\
\rowcolor{mycyan} GCPA  \cite{gcpa}    & R-50  & ss    & .864 & .102 & .869 & .064    & .945 & .037 & .933 & .032  & .886 & .038   & .801     & .056 \\
\rowcolor{mycyan}         & R-50  & ms    & .867 & .100 & .874  & .064   & .945 & .038 & .936 & .033  & .892 & .038   & .812     & .056 \\\hline
         & R-50  & orig. & .876 & .094 & .872 & .065    & .946 & .035 & .935 & .030  & .885 & .040   & .821     & .059 \\
ITSD  \cite{itsd}    & R-50  & ss    & .875 & .093 & .871  & .061 & .943 & .034 & .933  & .028 & .880 & .039    & .809    & .058  \\
         & R-50  & ms    & .875 & .101 & .866 & .064    & .946 & .034 & .933 & .030  & .886 & .037   & .816     & .055 \\\hline
\rowcolor{mycyan}         & R-50  & orig. & .879 & .092  & .867 & .064    & .947  & .033 & .935 & .029  & .884  & .037  & .810 & .056  \\
\rowcolor{mycyan} MINet \cite{minet}   & R-50  & ss    & .867 & .101  & .873 & .062    & .940 & .038  & .929 & .032  & .881 & .037   & .802   & .054   \\
\rowcolor{mycyan} & R-50  & ms    & .871 & .098  & .874 & .064    & .945 & .037  & .935 & .030  & .890 & .036   & .819     & .053 \\

\noalign{\smallskip}\bottomrule
\end{tabular}
\end{table*}

\section{The Proposed Loss Function}
\label{sec:loss}
Existing loss functions supervise the SOD methods from two perspectives: local supervision (BCE and CTLoss) and global supervision (Dice, $F_{\beta}$ and so on).
Local supervision calculates gradients for each pixel based on the predictions in a limited neighbor region.
For example, CTLoss outperforms the BCE loss on the max-$F_{\beta}$ score by introducing contour cues as weight maps of training samples.
However, the networks obtain a low ave-$F_{\beta}$ score when image edges are mistaken.
On the other hand, global supervision reduces the difference in global statistics between the predicted maps and the ground truths.
For instance, $F_{\beta}$ loss \cite{Floss} maximizes the $F_{\beta}$ scores between these two maps.
In practice, boundary pixels of salient objects are often rare than the other pixels within an image.
The network trained by $F_{\beta}$ loss will pay more attention on non-boundary pixels.
Thus, this loss can significantly increase the ave-$F_{\beta}$ score, but reduces the accuracy of the boundary pixels, resulting in a low max-$F_{\beta}$ score.
Although $F_{\beta}$ and CTLoss have their own shortages, they are complementary to each other.
Specifically, $F_{\beta}$ loss can assist CTLoss to distinguish the salient objects as well as their contours.
Meanwhile, since contour pixels are hard examples in SOD task, we can weight gradients in global supervision to force the network to emphasis on object contours.

Based on the above analysis, an Edge-Aware (EA) loss is developed to integrate more contour information.
The proposed loss is a combination of two terms, including CTLoss \cite{contour} and a new $L_{fc}$ loss:
\begin{equation}
L_{ea} = L_{c} + \lambda L_{fc},
\end{equation}
where $L_{c}$ is the contour loss as shown in Eqn. \ref{equ:lf}.
The $L_{fc}$ is developed by extending the contour cues as weight maps.
The contour map is generated by:
\begin{equation}
M = max\_pool(Y) \cdot max\_pool(1-Y),
\end{equation}
where, $M$ and $Y$ are contour and saliency masks respectively.
We use max pooling function with $3 \times 3$ kernel and $1 \times 1$ stride to simulate the dilation operation.
The contour maps are generated by calculating the product of dilated positive and negative maps.
Using these contour maps as weights, the weighted TP map is:
\begin{equation}
TP_{m} = M \cdot TP,
\end{equation}
where $m$ means the dot product with contour masks $M$.
Meanwhile, TP can be popularized to other maps similarly.
Subsequently, the weighted $F_{\beta}$ loss can be described as:
\begin{align}
L_{fc} &= 1 - \frac{(1+\beta^{2}) TP_{m}}{D_{m}}, \\
D_{m} &= (TN + TP)_{m} + \beta^{2}\times(FP + TP)_{m}.
\end{align}
The gradients of $L_{fc}$ can be computed by:
\begin{equation}
\frac{\partial L_{fc}}{\partial x_{i}} = m_{i} \times ( \frac{(1+\beta^{2}) TP_{m}}{D_{m}^{2}}- \frac{(1+\beta^{2})y_{i}}{D_{m}}),
\end{equation}
where $m_{i}$ and $y_{i}$ represent ground truth values in the contour and saliency maps respectively.

\section{Benchmark Experiments}
The proposed benchmark is implemented using PyTorch \cite{Pytorch} and run on a NVIDIA GeForce GTX 1080 Ti.
All models are trained with the consistent settings elaborated in Tab. \ref{tab:setting} and Section \ref{sec:setting}.
For optimization, we use a batch size of 8 for 20 epochs and adjust the learning rates independently to better explore the potential of these methods.

\begin{table*}
\renewcommand\arraystretch{1.1}
\caption{Quantitative results on the proposed SALOD  benchmark.
All methods are trained by the same settings in our benchmark, including backbones, input size and data augmentation strategies.
FPS evaluates the inference time for each method.
\textcolor{red}{Red}, \textcolor{blue}{Blue} and \textcolor{green}{Green} text indicate the best, second and third best performance respectively.
}
\centering
\footnotesize
\label{tab:benchmark}
\begin{tabular}{p{1.5cm}|p{1.5cm}<{\centering}p{1.1cm}<{\centering}p{1.0cm}<{\centering}|p{1.2cm}<{\centering}p{1.1cm}<{\centering}p{1.1cm}<{\centering}p{1.1cm}<{\centering}p{1.0cm}<{\centering}p{1.0cm}<{\centering}}
\toprule
Methods    & $\#$ Para.(M)  & MACs(G)  & FPS   & max-$F_\beta$ $\uparrow$ & ave-$F_\beta$ $\uparrow$ & Fbw $\uparrow$   & MAE $\downarrow$   & SM $\uparrow$   & EM $\uparrow$   \\
\midrule
\multicolumn{9}{l}{VGG16-based}                                                         \\\hline
DHSNet  \cite{dhsnet}  & \textcolor{red}{15.4} & \textcolor{blue}{52.5}   & \textcolor{red}{69.8} & .884  & .815  & .812  & .049  & .880 & .893      \\
Amulet  \cite{amulet}  & 33.2 & 1362   & 35.1 & .855  & .790  & .772  & .061  & .854 & .876      \\
NLDF    \cite{nldf}    & 24.6 & 136    & \textcolor{green}{46.3} & .886  & .824  & .828  & .045  & .881 & .898      \\
SRM     \cite{srm}     & 37.9 & \textcolor{green}{73.1}   & \textcolor{blue}{63.1} & .857  & .779  & .769  & .060  & .859 & .874      \\
PicaNet \cite{picanet} & 26.3 & 74.2   & 8.8  & .889  & .819  & .823  & .046  & .884 & .899      \\
DSS     \cite{dss}     & 62.2 & 99.4   & 30.3 & .891  & .827  & .826  & .046  & .888 & .899      \\
BASNet  \cite{basnet}  & 80.5 & 114.3  & 32.6  & \textcolor{blue}{.906}  & \textcolor{red}{.853}  & \textcolor{red}{.869}  & \textcolor{red}{.036}  & \textcolor{green}{.899} & \textcolor{red}{.915}      \\
CPD     \cite{cpd}     & 29.2 & 85.9   & 36.3  & .886  & .815  & .792  & .052  & .885 & .888      \\
PoolNet \cite{poolnet} & 52.5 & 236.2  & 23.1  & .902  & \textcolor{blue}{.850}  & \textcolor{green}{.852}  & \textcolor{green}{.039}  & .898 & \textcolor{green}{.913}      \\
EGNet   \cite{egnet}   & 101  & 178.8  & 16.3  & \textcolor{red}{.909}  & \textcolor{red}{.853}  & \textcolor{blue}{.859}  & \textcolor{blue}{.037}  & \textcolor{red}{.904} & \textcolor{blue}{.914}      \\
SCRN    \cite{scrn}    & \textcolor{blue}{16.3} & \textcolor{red}{47.2}   & 24.8  & .896  & .820  & .822  & .046  & .891 & .894      \\
GCPA    \cite{gcpa}    & 42.8 & 197.1  & 29.3  & .903  & .836  & .845  & .041  & .898 & .907      \\
ITSD    \cite{itsd}    & \textcolor{green}{16.9} & 76.3   & 30.6  & \textcolor{green}{.905}  & .820  & .834  & .045  & \textcolor{blue}{.901} & .896      \\
MINet   \cite{minet}   & 47.8 & 162    & 23.4  & .900  & \textcolor{green}{.839}  & \textcolor{green}{.852}  & \textcolor{green}{.039}  & .895 & .909      \\\hline
\multicolumn{9}{l}{ResNet50-based}                                                      \\\hline
DHSNet  \cite{dhsnet}  & \textcolor{red}{24.2} & \textcolor{blue}{13.8}   & \textcolor{red}{49.2}  & .909  & .830  & .848  & .039  & .905 & .905      \\
Amulet  \cite{amulet}  & 79.8 & 1093.8 & \textcolor{green}{35.1}  & .895  & .822  & .835  & .042  & .894 & .900      \\
NLDF    \cite{nldf}    & 41.1 & 115.1  & 30.5  & .903  & .837  & .855  & .038  & .898 & .910      \\
SRM     \cite{srm}     & 61.2 & 20.2   & 34.3  & .882  & .803  & .812  & .047  & .885 & .891      \\
PicaNet \cite{picanet} & 106.1 & 36.9  & 14.8  & .904  & .823  & .843  & .041  & .902 & .902      \\
DSS     \cite{dss}     & 134.3 & 35.3  & 27.3  & .894  & .821  & .826  & .045  & .893 & .898      \\
BASNet  \cite{basnet}  & 95.5 & 47.2   & 32.8  & \textcolor{red}{.917}  & \textcolor{red}{.861}  & \textcolor{red}{.884}  & \textcolor{red}{.032}  & \textcolor{blue}{.909} & \textcolor{red}{.921}      \\
CPD     \cite{cpd}     & 47.9 & \textcolor{green}{14.7}   & 22.7  & .906  & .842  & .836  & .040  & .904 & .908      \\
PoolNet \cite{poolnet} & 68.3 & 66.9   & 33.9  & .912  & \textcolor{green}{.843}  & .861  & .036  & \textcolor{green}{.907} & .912      \\
EGNet   \cite{egnet}   & 111.7 & 222.8 & 10.2  & \textcolor{red}{.917}  & \textcolor{blue}{.851}  & \textcolor{green}{.867}  & .036  & \textcolor{red}{.912} & \textcolor{green}{.914}      \\
SCRN    \cite{scrn}    & \textcolor{blue}{25.2}  & \textcolor{red}{12.5}  & 19.3  & .910  & .838  & .845  & .040  & .906 & .905      \\
GCPA    \cite{gcpa}    & 67.1  & 54.3  & \textcolor{blue}{37.8}  & \textcolor{blue}{.916}  & .841  & .866  & \textcolor{green}{.035}  & \textcolor{red}{.912} & .912      \\
ITSD    \cite{itsd}    & \textcolor{green}{25.7}   & 19.6 & 29.4  & \textcolor{green}{.913}  & .825  & .842  & .042  & \textcolor{green}{.907} & .899      \\
MINet   \cite{minet}   & 162.4  & 87   & 23.5  & \textcolor{green}{.913}  & \textcolor{blue}{.851}  & \textcolor{blue}{.871}  & \textcolor{blue}{.034}  & .906 & \textcolor{blue}{.917}      \\
\noalign{\smallskip}\bottomrule
\end{tabular}
\end{table*}

\begin{figure*}[!t]
\centering

\begin{minipage}{1 \textwidth}
\begin{minipage}{0.1 \textwidth} Image \vspace{1.5cm} \end{minipage}
\includegraphics[width=0.9in,height=0.7in]{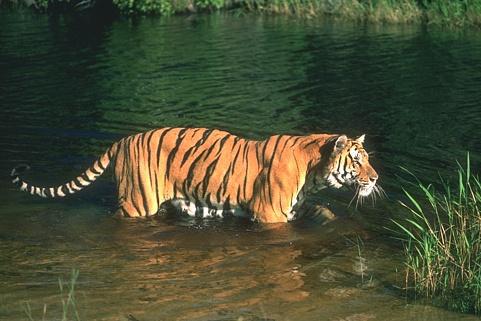}
\includegraphics[width=0.9in,height=0.7in]{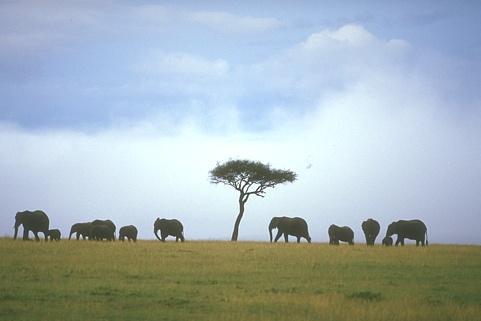}
\includegraphics[width=0.9in,height=0.7in]{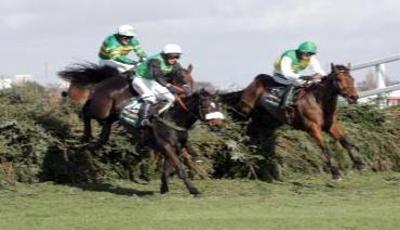}
\includegraphics[width=0.9in,height=0.7in]{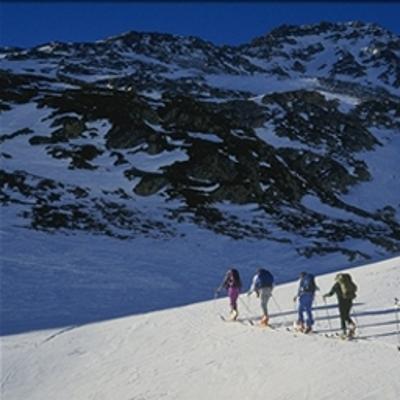}
\includegraphics[width=0.9in,height=0.7in]{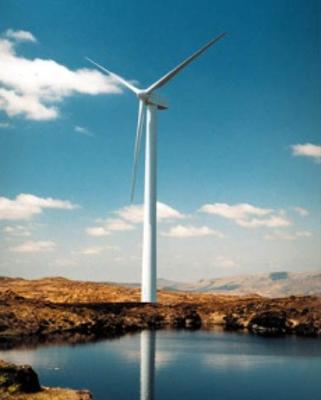}
\includegraphics[width=0.9in,height=0.7in]{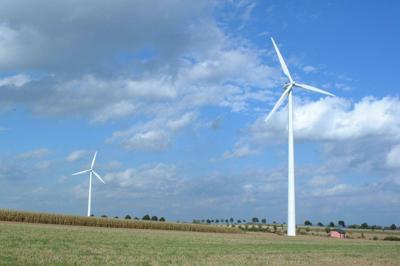}
\end{minipage}
\vspace{-0.8cm}

\begin{minipage}{1 \textwidth}
\begin{minipage}{0.1 \textwidth} GT \vspace{1.5cm} \end{minipage}
\includegraphics[width=0.9in,height=0.7in]{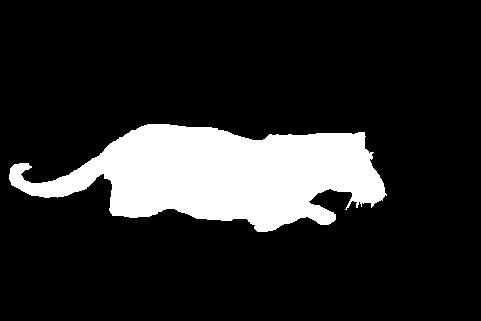}
\includegraphics[width=0.9in,height=0.7in]{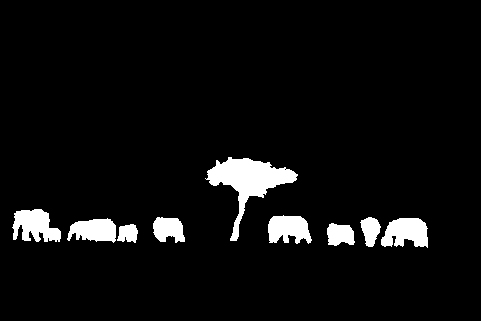}
\includegraphics[width=0.9in,height=0.7in]{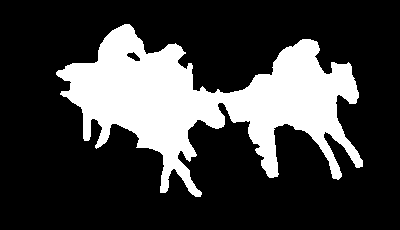}
\includegraphics[width=0.9in,height=0.7in]{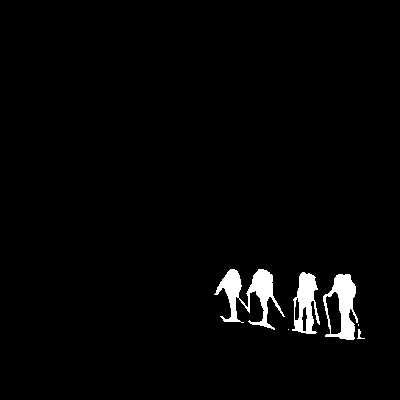}
\includegraphics[width=0.9in,height=0.7in]{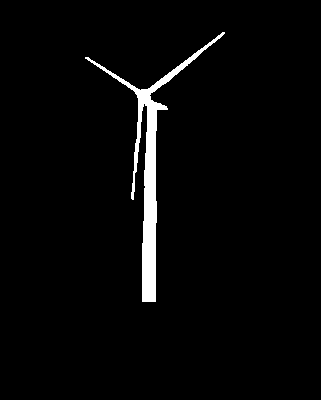}
\includegraphics[width=0.9in,height=0.7in]{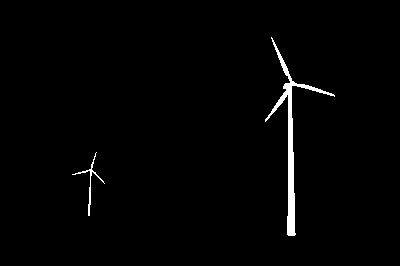}
\end{minipage}
\vspace{-0.8cm}

\begin{minipage}{1 \textwidth}
\begin{minipage}{0.1 \textwidth} DHSNet \vspace{1.5cm} \end{minipage}
\includegraphics[width=0.9in,height=0.7in]{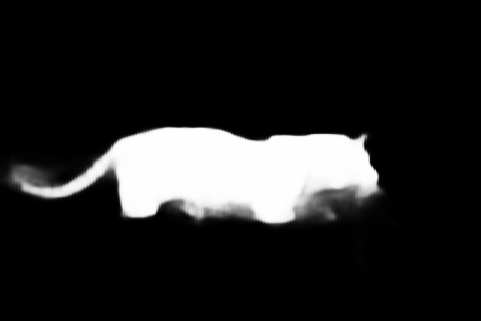}
\includegraphics[width=0.9in,height=0.7in]{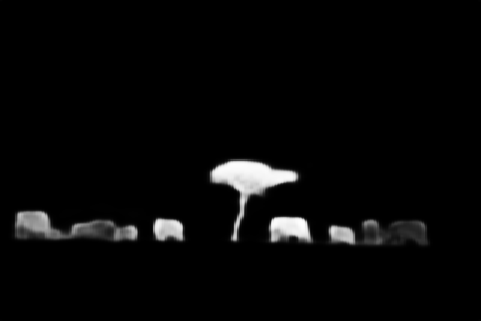}
\includegraphics[width=0.9in,height=0.7in]{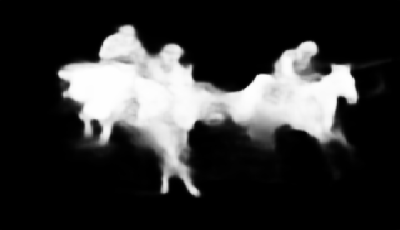}
\includegraphics[width=0.9in,height=0.7in]{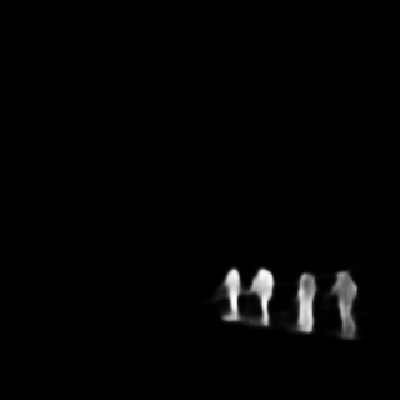}
\includegraphics[width=0.9in,height=0.7in]{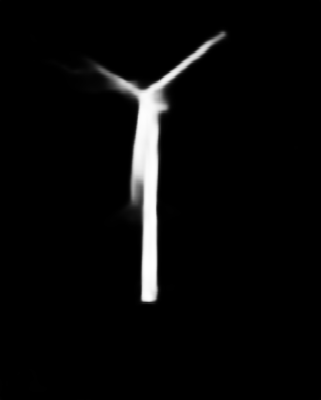}
\includegraphics[width=0.9in,height=0.7in]{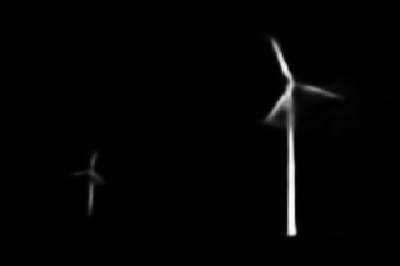}
\end{minipage}
\vspace{-0.8cm}

\begin{minipage}{1 \textwidth}
\begin{minipage}{0.1 \textwidth} Amulet \vspace{1.5cm} \end{minipage}
\includegraphics[width=0.9in,height=0.7in]{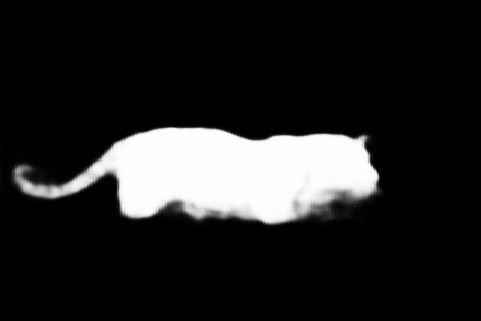}
\includegraphics[width=0.9in,height=0.7in]{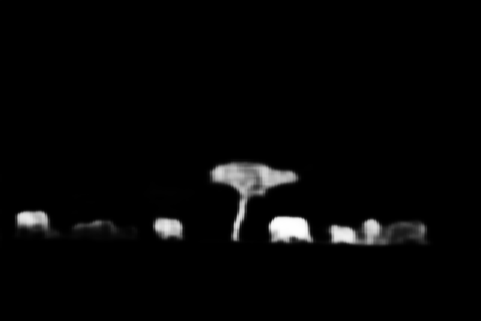}
\includegraphics[width=0.9in,height=0.7in]{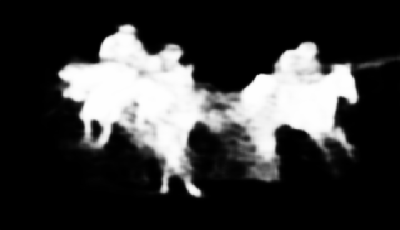}
\includegraphics[width=0.9in,height=0.7in]{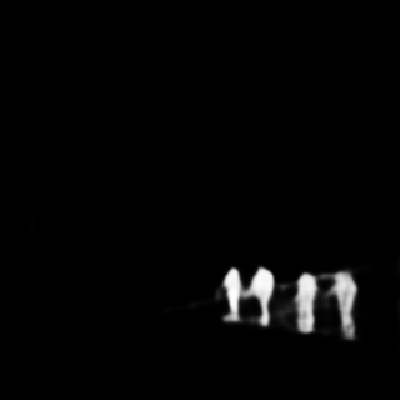}
\includegraphics[width=0.9in,height=0.7in]{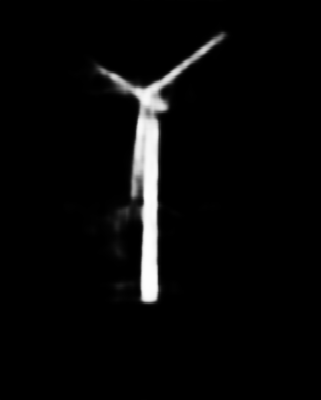}
\includegraphics[width=0.9in,height=0.7in]{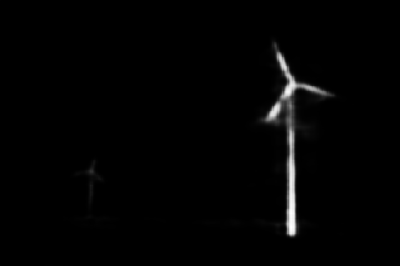}
\end{minipage}
\vspace{-0.8cm}

\begin{minipage}{1 \textwidth}
\begin{minipage}{0.1 \textwidth} SRM \vspace{1.5cm} \end{minipage}
\includegraphics[width=0.9in,height=0.7in]{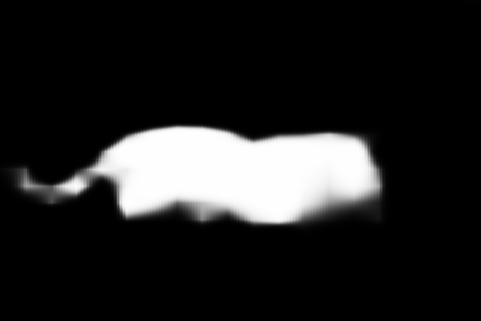}
\includegraphics[width=0.9in,height=0.7in]{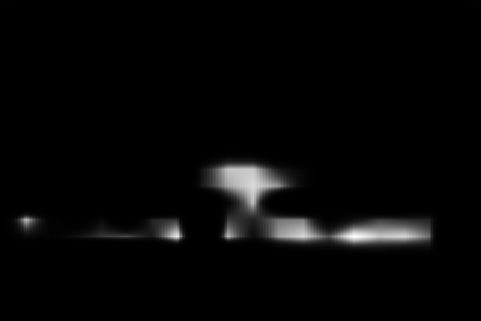}
\includegraphics[width=0.9in,height=0.7in]{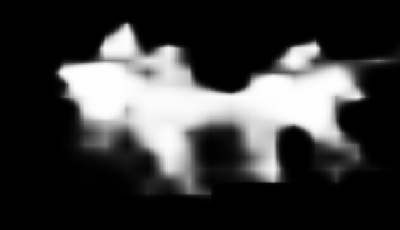}
\includegraphics[width=0.9in,height=0.7in]{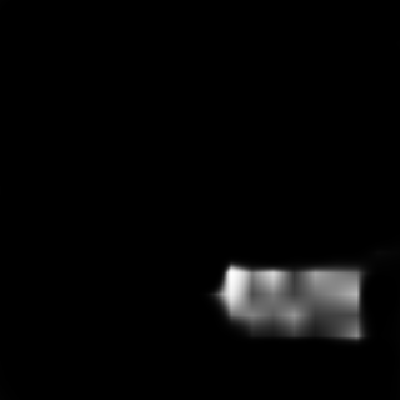}
\includegraphics[width=0.9in,height=0.7in]{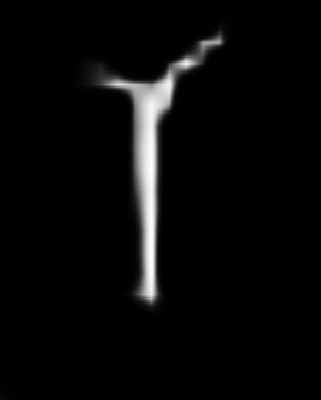}
\includegraphics[width=0.9in,height=0.7in]{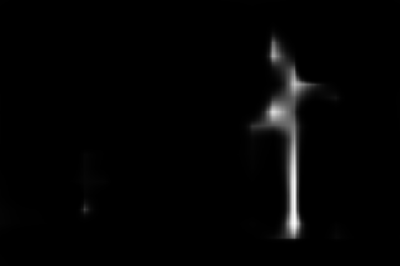}
\end{minipage}
\vspace{-0.8cm}

\begin{minipage}{1 \textwidth}
\begin{minipage}{0.1 \textwidth} DSS \vspace{1.5cm} \end{minipage}
\includegraphics[width=0.9in,height=0.7in]{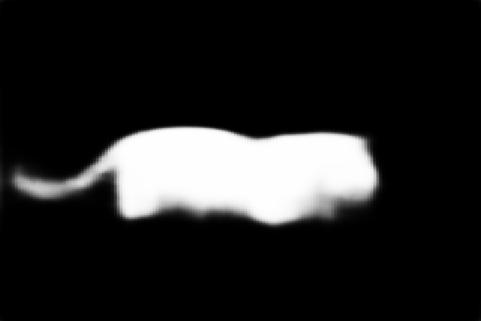}
\includegraphics[width=0.9in,height=0.7in]{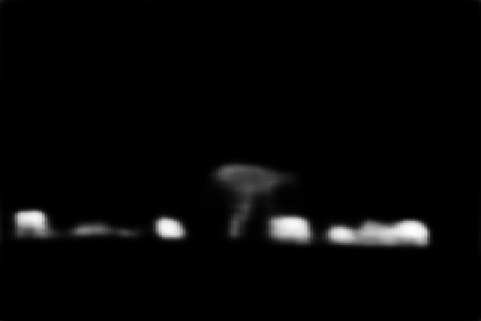}
\includegraphics[width=0.9in,height=0.7in]{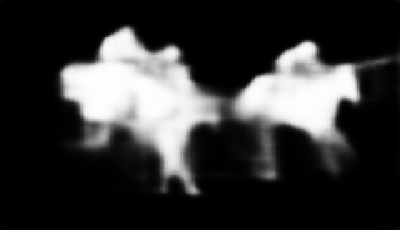}
\includegraphics[width=0.9in,height=0.7in]{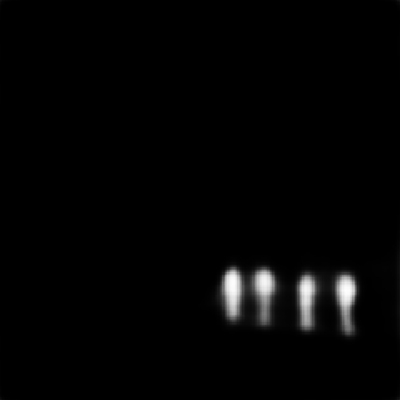}
\includegraphics[width=0.9in,height=0.7in]{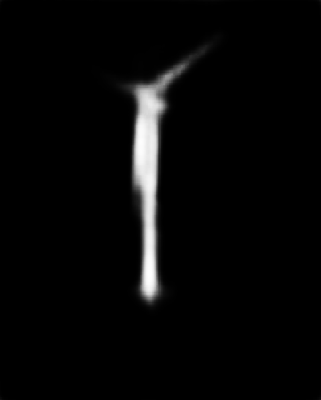}
\includegraphics[width=0.9in,height=0.7in]{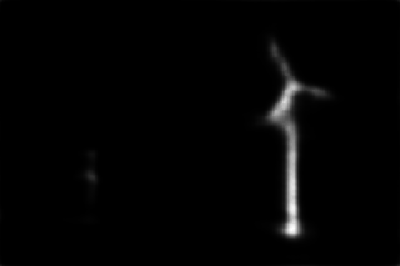}
\end{minipage}
\vspace{-0.8cm}

\begin{minipage}{1 \textwidth}
\begin{minipage}{0.1 \textwidth} BASNet \vspace{1.5cm} \end{minipage}
\includegraphics[width=0.9in,height=0.7in]{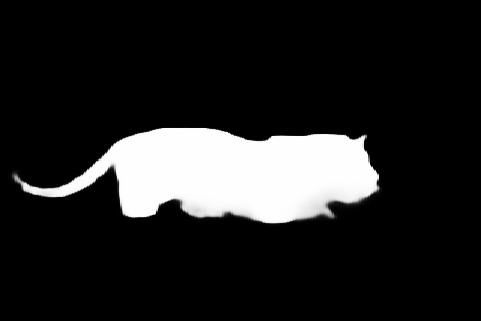}
\includegraphics[width=0.9in,height=0.7in]{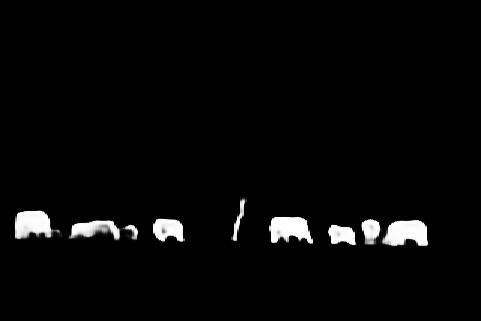}
\includegraphics[width=0.9in,height=0.7in]{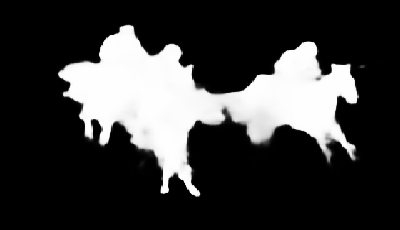}
\includegraphics[width=0.9in,height=0.7in]{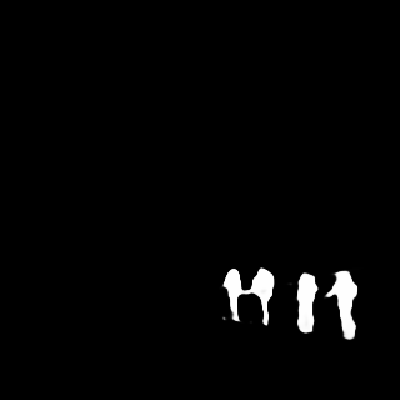}
\includegraphics[width=0.9in,height=0.7in]{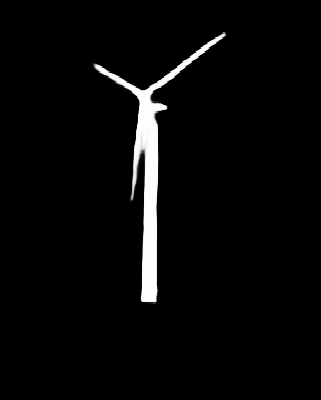}
\includegraphics[width=0.9in,height=0.7in]{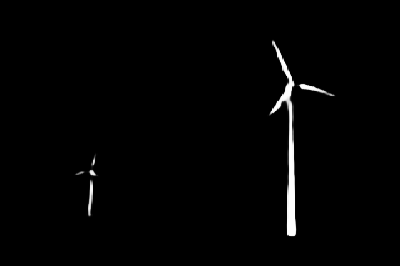}
\end{minipage}
\vspace{-0.8cm}

\begin{minipage}{1 \textwidth}
\begin{minipage}{0.1 \textwidth} CPD \vspace{1.5cm} \end{minipage}
\includegraphics[width=0.9in,height=0.7in]{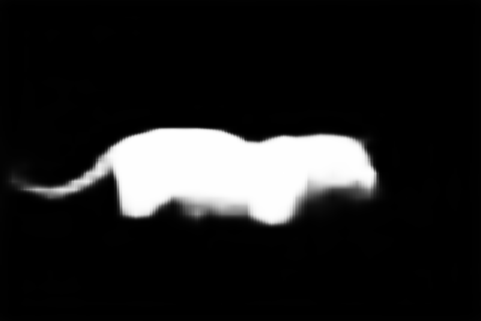}
\includegraphics[width=0.9in,height=0.7in]{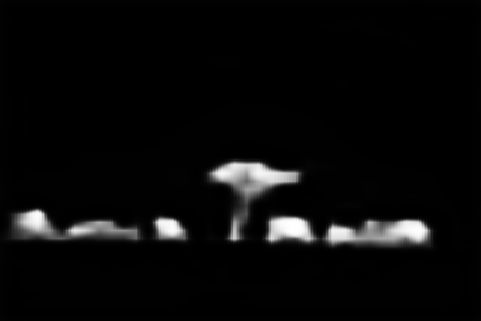}
\includegraphics[width=0.9in,height=0.7in]{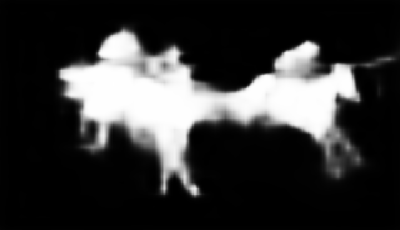}
\includegraphics[width=0.9in,height=0.7in]{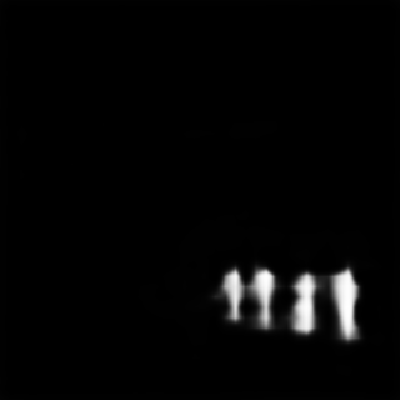}
\includegraphics[width=0.9in,height=0.7in]{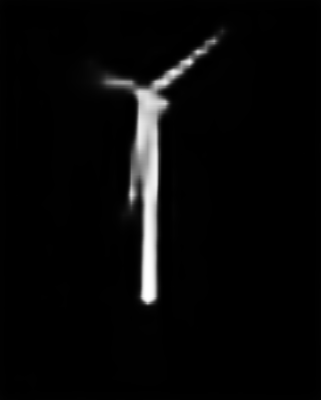}
\includegraphics[width=0.9in,height=0.7in]{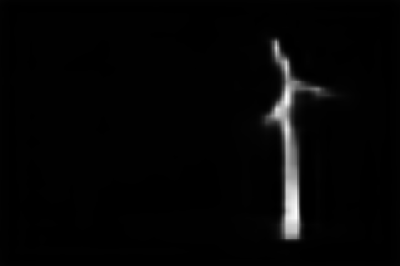}
\end{minipage}
\vspace{-0.8cm}

\begin{minipage}{1 \textwidth}
\begin{minipage}{0.1 \textwidth} PoolNet \vspace{1.5cm} \end{minipage}
\includegraphics[width=0.9in,height=0.7in]{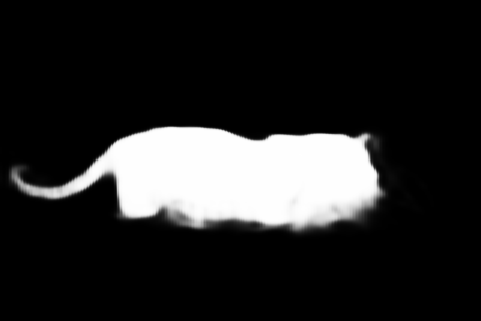}
\includegraphics[width=0.9in,height=0.7in]{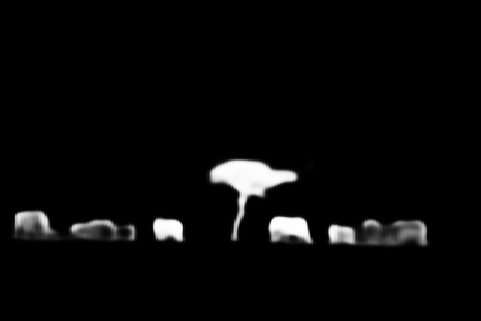}
\includegraphics[width=0.9in,height=0.7in]{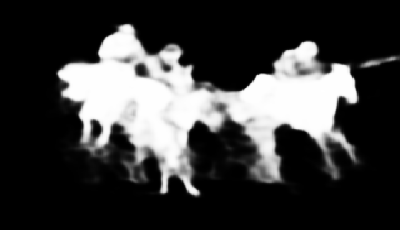}
\includegraphics[width=0.9in,height=0.7in]{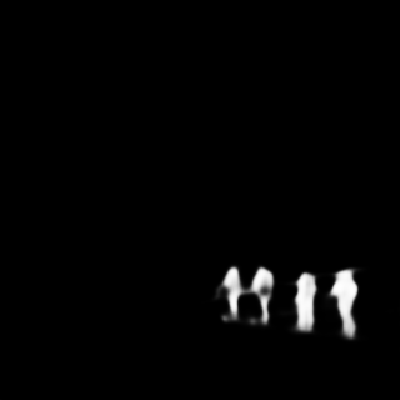}
\includegraphics[width=0.9in,height=0.7in]{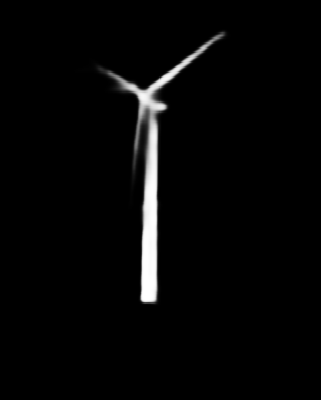}
\includegraphics[width=0.9in,height=0.7in]{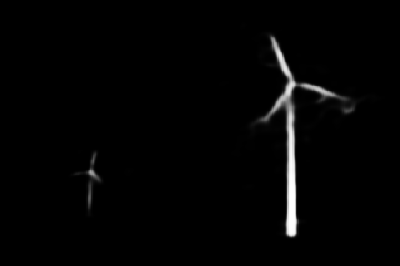}
\end{minipage}
\vspace{-0.8cm}

\begin{minipage}{1 \textwidth}
\begin{minipage}{0.1 \textwidth} ITSD \vspace{1.5cm} \end{minipage}
\includegraphics[width=0.9in,height=0.7in]{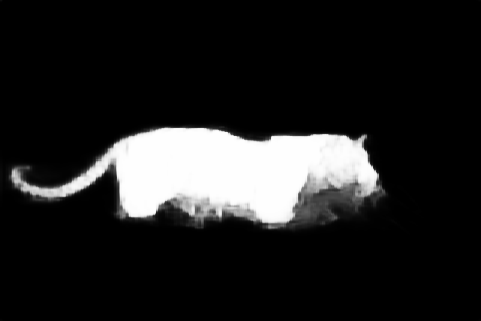}
\includegraphics[width=0.9in,height=0.7in]{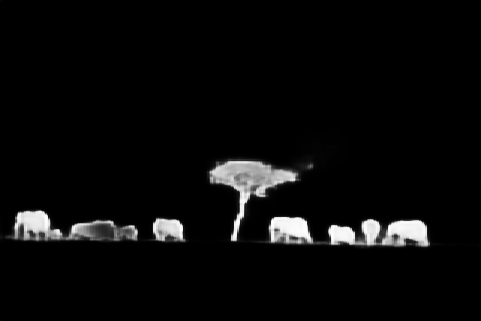}
\includegraphics[width=0.9in,height=0.7in]{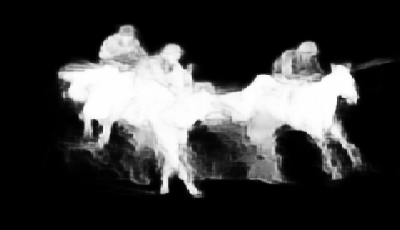}
\includegraphics[width=0.9in,height=0.7in]{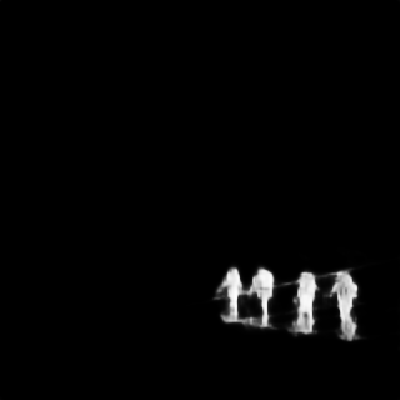}
\includegraphics[width=0.9in,height=0.7in]{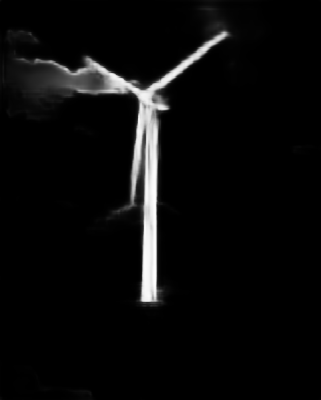}
\includegraphics[width=0.9in,height=0.7in]{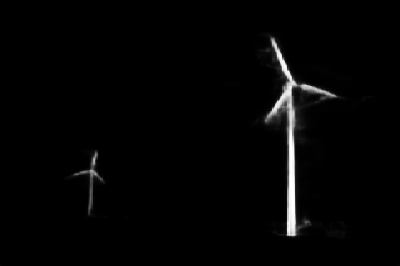}
\end{minipage}
\vspace{-0.8cm}

\begin{minipage}{1 \textwidth}
\begin{minipage}{0.1 \textwidth} MINet \vspace{1.5cm} \end{minipage}
\includegraphics[width=0.9in,height=0.7in]{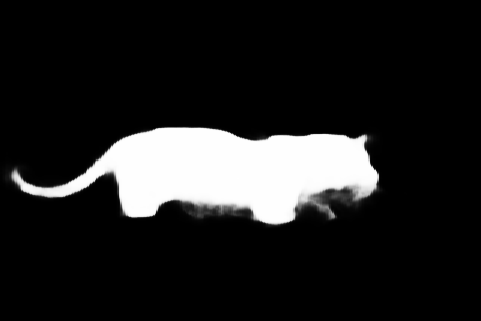}
\includegraphics[width=0.9in,height=0.7in]{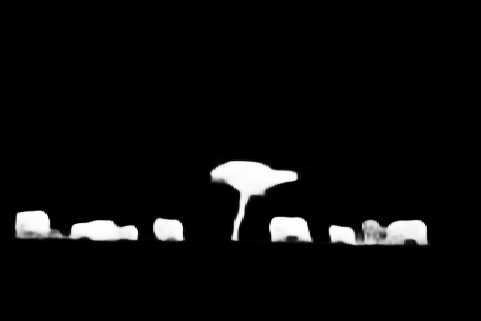}
\includegraphics[width=0.9in,height=0.7in]{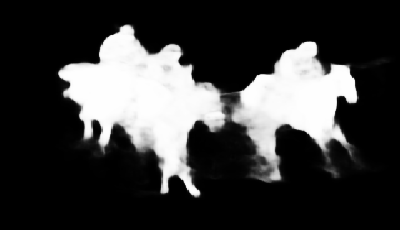}
\includegraphics[width=0.9in,height=0.7in]{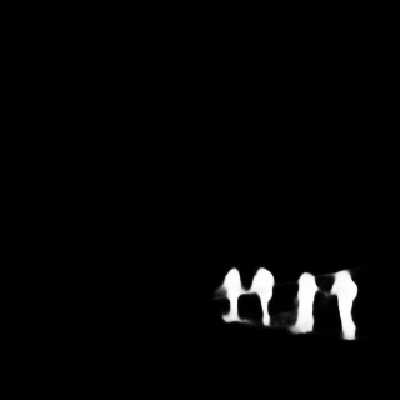}
\includegraphics[width=0.9in,height=0.7in]{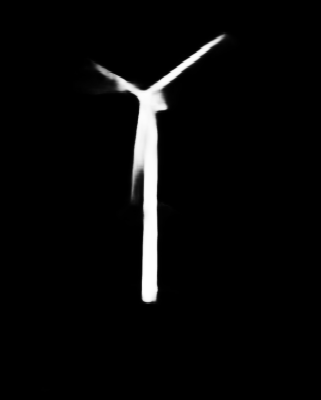}
\includegraphics[width=0.9in,height=0.7in]{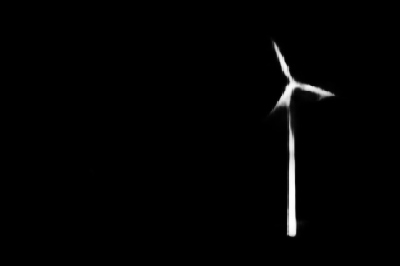}
\end{minipage}
\vspace{-0.8cm}

\caption{Example predictions of several representative SOD models.
}
\label{fig:corr}
\end{figure*}

\subsection{Preliminary Experiment}
We first conduct a preliminary experiment to verify that our re-implementations are comparable with their original implementations.
To avoid the performance discrepancy caused by different evaluation processes, we collect the predictions from official implementations and compute their scores using the same evaluation code.
For our re-implementations, we train them on the DUTS-TR \cite{DUTS} dataset and employ six test sets to validate the effectiveness, including SOD \cite{SOD}, PASCAL-S \cite{PASCAL-S}, ECSSD \cite{ecssd}, HKU-IS \cite{HKU-IS}, DUTS-TE \cite{DUTS} and DUT-OMRON \cite{DUT-OMRON}.
In addition, the multi-scale training strategy is adopted since it is also widely used in many existing SOD methods, such as SCRN \cite{scrn} and ITSD \cite{itsd}.
All results are exhibited in Tab. \ref{tab:reproduce}.

Experiment results prove that our re-implementations achieve similar results compared with their original ones.
Specifically, some earlier works obtain substantial improvements when training on DUTS-TR \cite{DUTS} or upgrading backbones to ResNet-50, such as DHSNet \cite{dhsnet}, Amulet \cite{amulet} and NLDF \cite{nldf}.
An interesting finding is that BASNet \cite{basnet} with ResNet-50 backbone achieves state-of-the-art performance among the available methods in our benchmark.
Moreover, several methods surpass their original implementations owing to the advanced optimization strategies, such as SRM \cite{srm}, DSS \cite{dss} and CPD \cite{cpd}.
Results shown in Tab. \ref{tab:reproduce} prove that some performance discrepancies are caused by different settings of SOD methods, such as input image resolutions.
For example, PicaNet \cite{picanet} attains better results when its input is expanded from $224 \times 224$ to $320 \times 320$.
In addition, multi-scale training strategy helps some methods to improve their performance, such as PoolNet \cite{poolnet} and EGNet \cite{egnet}.

\begin{figure*}[!t]
\centering
\includegraphics[width=6.0in,height=4.8in]{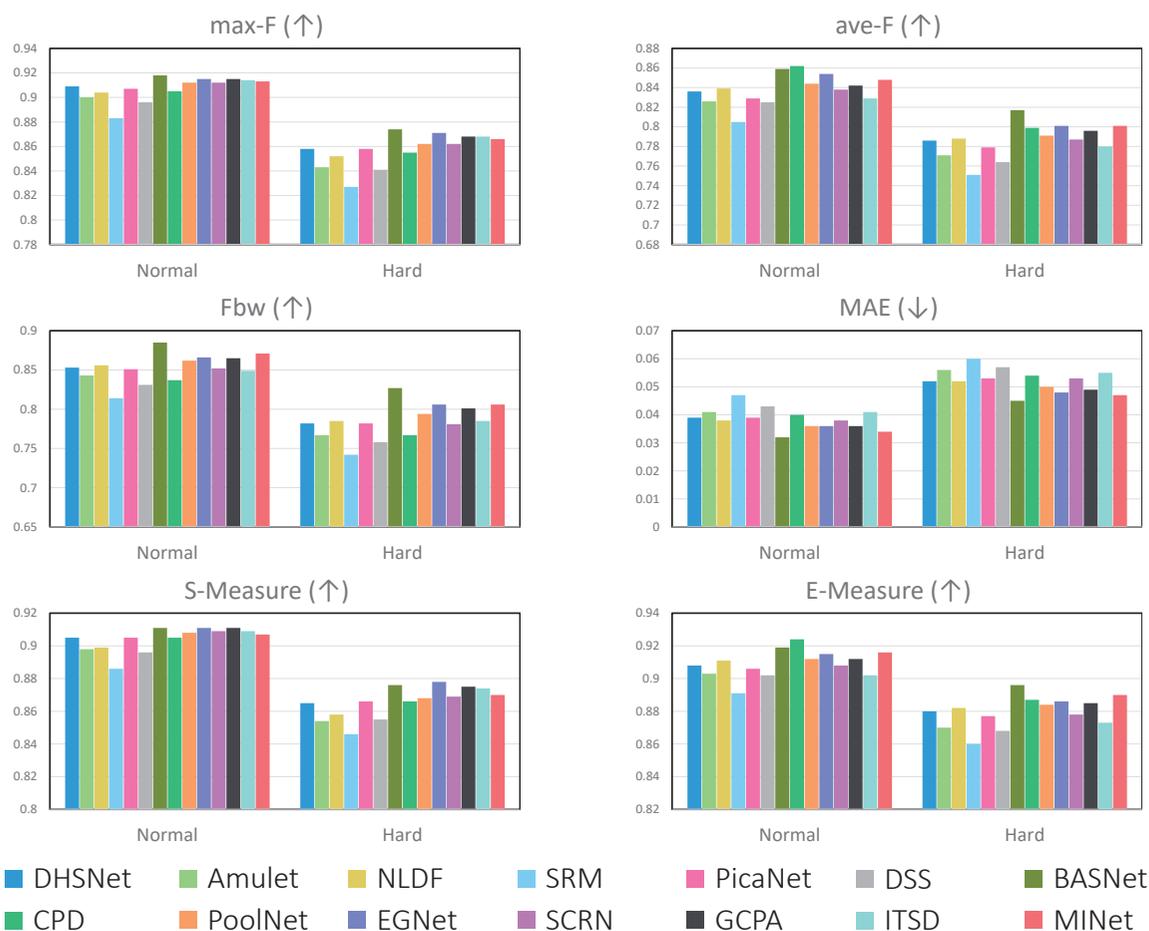}
\caption{ Results of objectness shifting experiment. All models are trained on the easy set of the proposed SALOD dataset.}
\label{fig:objectness}
\end{figure*}

\subsection{Benchmark Results}
In Tab. \ref{tab:benchmark}, we show the benchmark results on six evaluation metrics as well as other efficiency criterions, including the number of Parameter (\# Param.), Multiply Accumulate Operations (MACs) and Frame Per Second (FPS) during testing.
Our benchmarking results show different trends as concluded in previous survey \cite{wang_survey}.
First, latest U-shape networks \cite{basnet} \cite{itsd} \cite{minet} surpass other networks with different typologies (SRM \cite{srm} and CPD \cite{cpd}).
It proves that U-shape structure can better integrate multi-level information to capture more details by using skip-connections and progressive upsampling strategy.
Second, these methods with ResNet-50 backbone report much smaller performance variance than VGG-16 backbone.
VGG-16 is relatively shallow and simple, so it is hard to extract powerful hierarchical features like ResNet-50.
Therefore, sophisticated decoders can further integrate multi-level information to produce more robust saliency predictions.
Third, large models, such as MINet \cite{minet} and GCPA \cite{gcpa}, outperform lightweight ones (SCRN \cite{scrn} and ITSD \cite{itsd}) on most metrics.
Interestingly, they also report comparable detection speed to lightweight methods owing to current code optimization.
Fourth, the improvements from DHSNet \cite{dhsnet} to MINet \cite{minet} is not as dramatic as reported in previous works.
Fifth, BASNet achieves state-of-the-art performance on all metrics compared with other SOD methods.
It is noteworthy that our conclusion is different from the previous survey \cite{wang_survey}, which considers that EGNet \cite{egnet}, PoolNet \cite{poolnet} and SCRN \cite{scrn} are better than BASNet \cite{basnet}.
Their conclusion is based on the official implementations and ignores the large discrepancy caused by different settings of these methods.
The consistent settings in our benchmark ensure that all methods are compared with each other in a relatively fair situation and thus reveal the real effectiveness of these methods.
Last, other methods also obtain impressive results in our benchmarking experiments.
For example, DHSNet \cite{dhsnet} reports competitive results with the fastest inference speed and the fewest parameters.
ITSD \cite{itsd} and SCRN \cite{scrn} provide a good balance between efficiency and effectiveness.
The performances of EGNet \cite{egnet} and MINet \cite{minet} are comparable to BASNet \cite{basnet}, while their FPS are much lower.

From the visual examples in Fig. \ref{fig:corr}, we can see that DHSNet \cite{dhsnet}, Amulet \cite{amulet} and PoolNet \cite{poolnet} output low confidence regions for object boundaries.
In addition, SRM \cite{srm}, DSS \cite{dss} and CPD \cite{cpd} usually produce coarse predictions because they are hard to capture more details.
BASNet \cite{basnet} predicts more distinct saliency maps than other methods, but fails to capture some part of target objects.
For example, the crown of the tree is lost in the second example.
ITSD \cite{itsd} precisely captures details and boundaries, but segments some conspicuous edges in background simultaneously.
In addition, MINet \cite{minet} segments most of salient objects but fails to perceive some small objects.

\subsection{Objectness Shifting Experiment}
We train SOD methods using the second protocol to evaluate their robustness against objectness-shifting problem.
In this protocol, the proposed SALOD dataset is split into three subsets, including easy, normal and hard sets.
The easy subset is used for training, and the other two subsets are used for evaluation.
We list the results of 14 methods on two subsets in Fig. \ref{fig:objectness} and elaborate the differences between these two sets in Tab. \ref{tab:gap}.

Combining all results from two subsets, we observe that the performance of SOD methods are degraded on all metrics.
Moreover, the lastest sophisticated decoders better integrate multi-level cues from the encoder features and thus result in a more robust performance than earlier works.
Specifically, PoolNet \cite{poolnet} and BASNet \cite{basnet} have the lowest reduction degrees on max-$F_\beta$ scores.
In addition, BASNet also performs well on ave-$F_\beta$ and Fbw scores.
EGNet \cite{egnet} reports the lowest performance drops on MAE and S-Measure scores, because its sophisticated decoder well preserve the shapes of objects.
Surprisingly, the lowest E-Measure drop is obtained by DHSNet \cite{dhsnet}, the oldest method in our benchmark.
Since E-Measure focuses on local pixel matching information, the Recurrent Convolutional Layers in DHSNet assist the network to better learn the local context and obtain more stable E-Measure scores.
The aforementioned conclusions demonstrate that the objectness-shifting can significantly degrades the detection performance, and thus future efforts should pay more attention to tackle this problem.

\begin{table}[t]
\renewcommand\arraystretch{1.1}
\setlength\tabcolsep{3pt}
\caption{Performance drops of SOD methods in our objectness shifting experiments. We calculate the difference between the normal and hard subsets.}
\label{tab:gap}
\centering
\footnotesize
\begin{tabular}{l|cccccc}
\toprule
\multirow{2}{*}{Methods} &\multicolumn{6}{c}{Performance drops $\Delta$ ($\downarrow$)} \\\cline{2-7}
                      & max-$F_\beta$ & ave-$F_\beta$ & Fbw   & MAE   & SM   & EM        \\\midrule
DHSNet  \cite{dhsnet} & .051  & .050  & .071  & .013  & .040 & \textbf{.020}      \\
Amulet  \cite{amulet} & .057  & .055  & .076  & .015  & .044 & .033      \\
NLDF    \cite{nldf}   & .052  & .051  & .071  & .014  & .041 & .029      \\
SRM     \cite{srm}    & .056  & .054  & .072  & .013  & .040 & .031      \\
PicaNet \cite{picanet}& .049  & .050  & .069  & .014  & .039 & .029      \\
DSS     \cite{dss}    & .055  & .061  & .073  & .014  & .041 & .034      \\
BASNet  \cite{basnet} & \textbf{.044}  & \textbf{.042}  & \textbf{.058}  & .013  & .035 & .023      \\
CPD     \cite{cpd}    & .050  & .063  & .070  & .014  & .039 & .037      \\
PoolNet \cite{poolnet}& \textbf{.044}  & .053  & .068  & .014  & .042 & .028      \\
EGNet   \cite{egnet}  & .050  & .053  & .060  & \textbf{.012}  & \textbf{.033} & .029      \\
SCRN    \cite{scrn}   & .050  & .051  & .071  & .015  & .040 & .030      \\
GCPA    \cite{gcb}    & .047  & .046  & .064  & .013  & .036 & .027      \\
ITSD    \cite{itsd}   & .046  & .049  & .064  & .014  & .035 & .029      \\
MINet   \cite{minet}  & .047  & .047  & .065  & .013  & .037 & .026      \\
\noalign{\smallskip}\bottomrule
\end{tabular}
\end{table}

\begin{figure*}[!t]
\centering
\includegraphics[width=0.46 \textwidth]{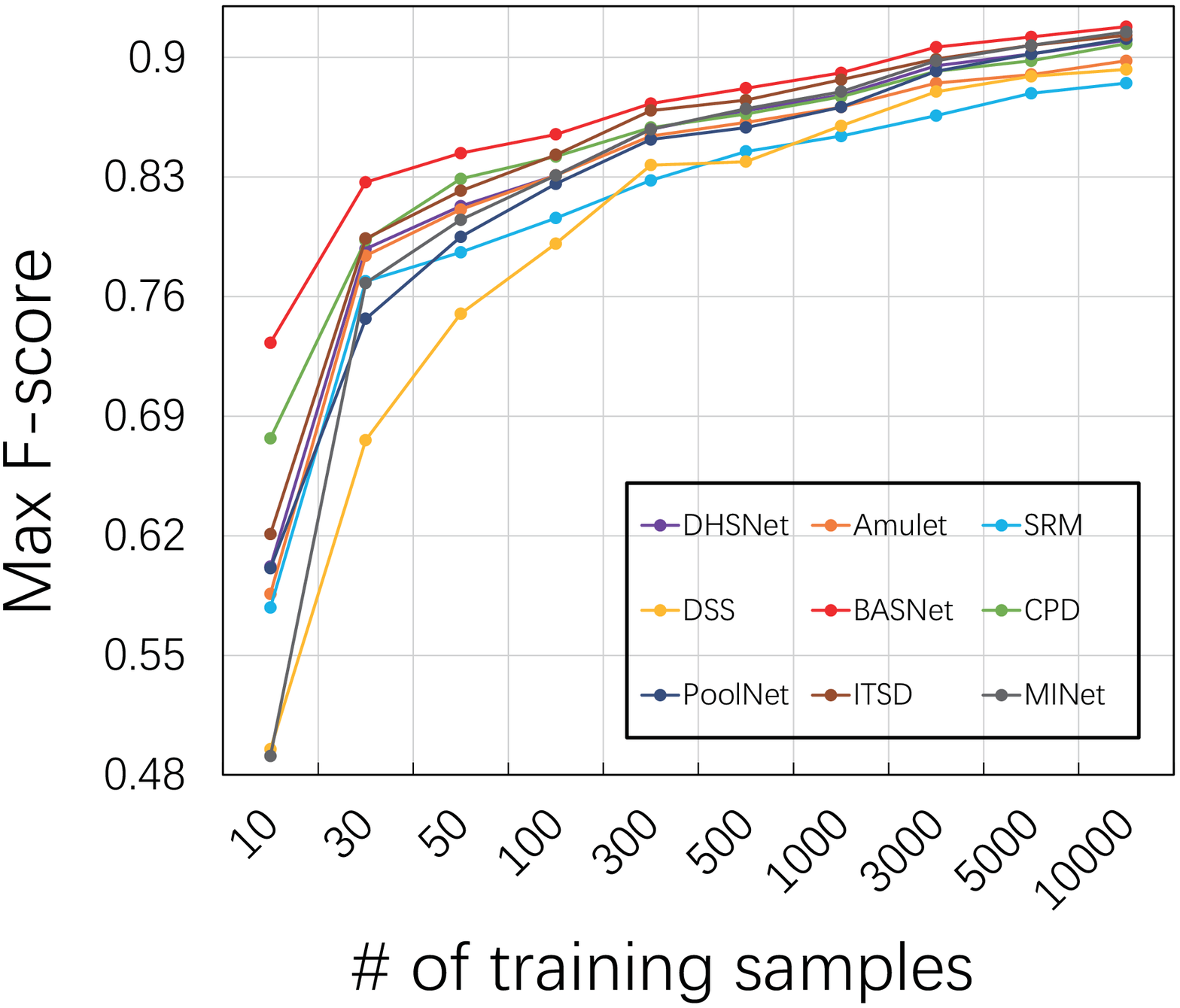}\hspace{20pt}
\includegraphics[width=0.46 \textwidth]{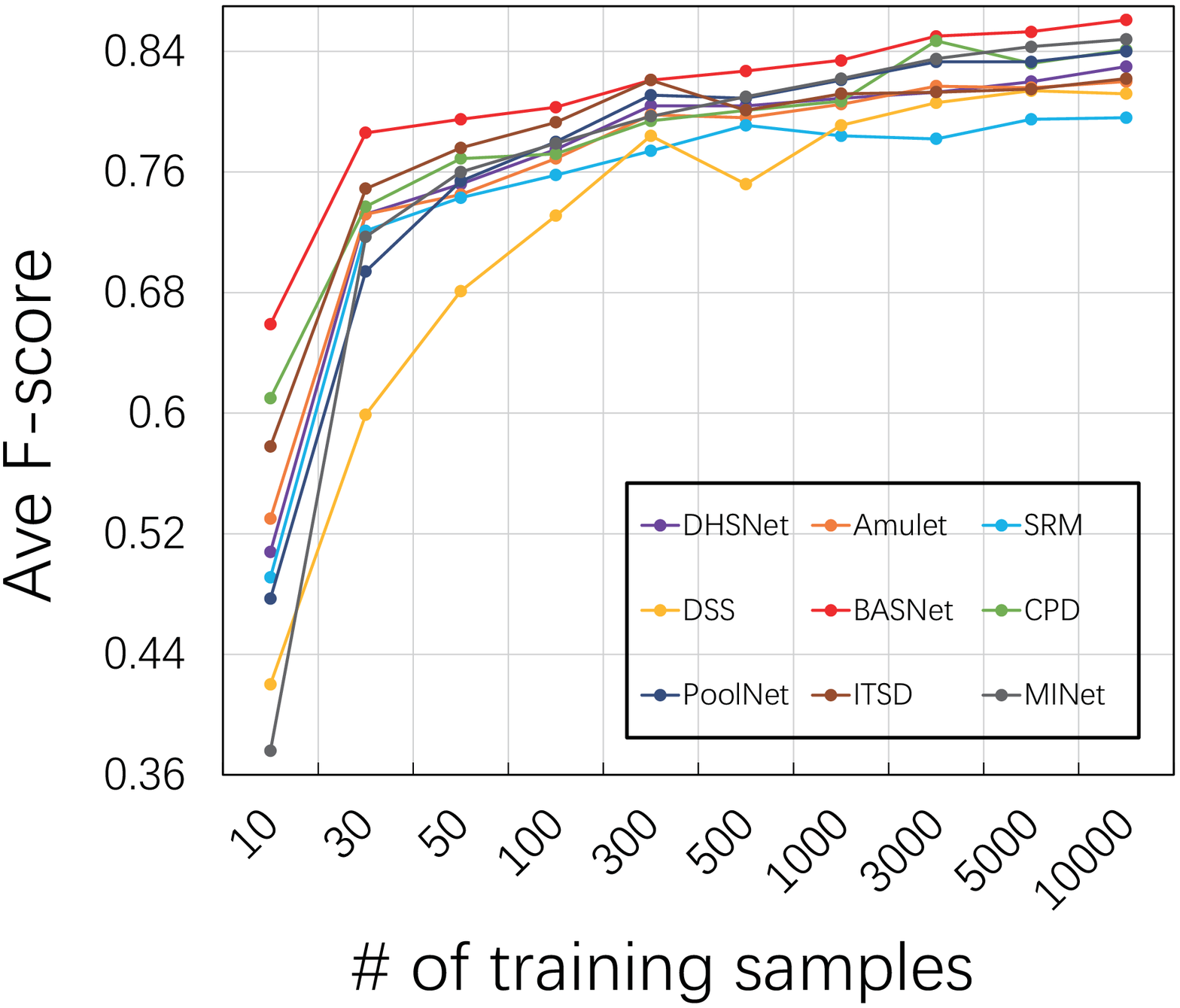}\vspace{15pt}
\includegraphics[width=0.46 \textwidth]{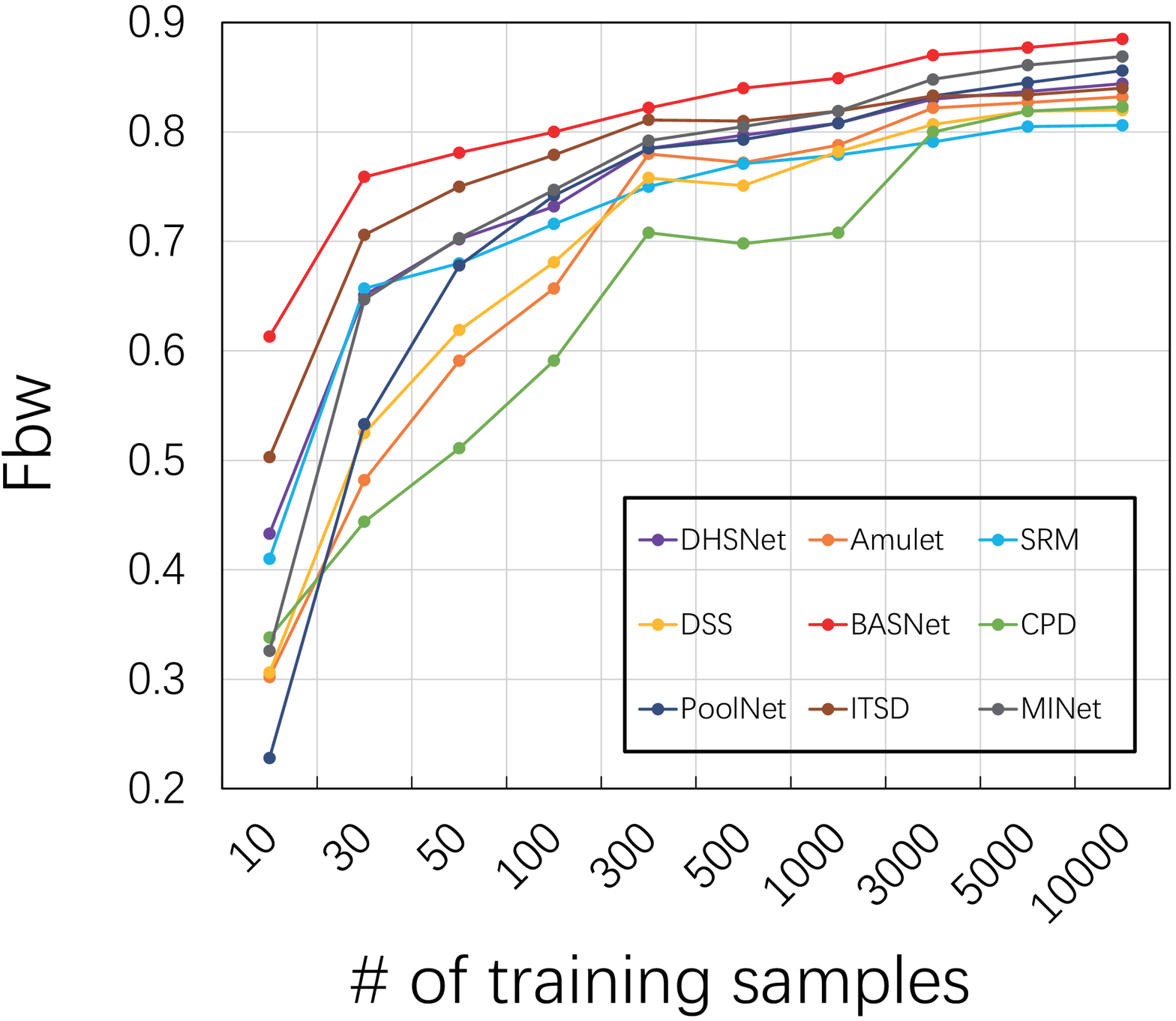}\hspace{20pt}
\includegraphics[width=0.46 \textwidth]{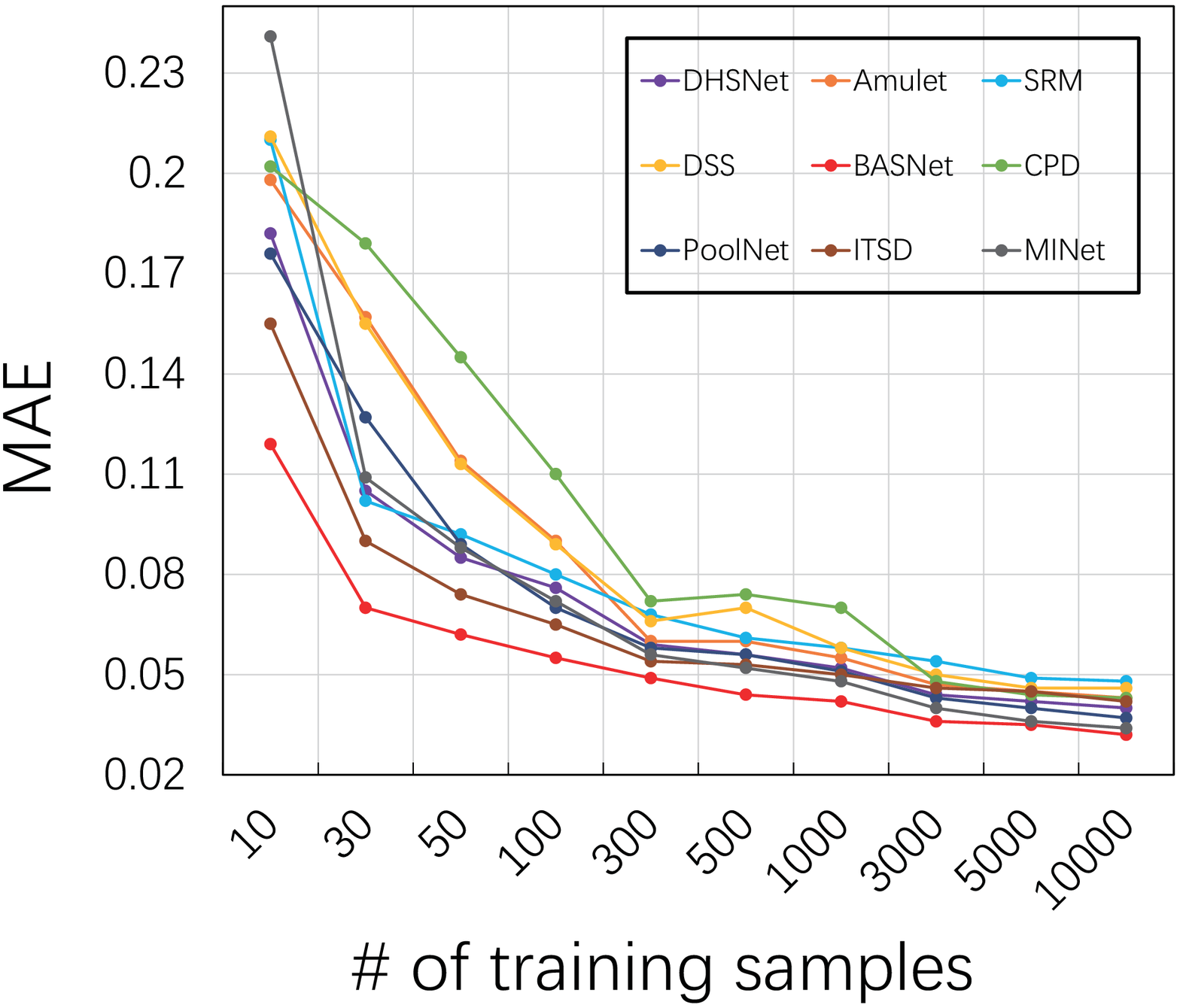}\vspace{15pt}
\includegraphics[width=0.46 \textwidth]{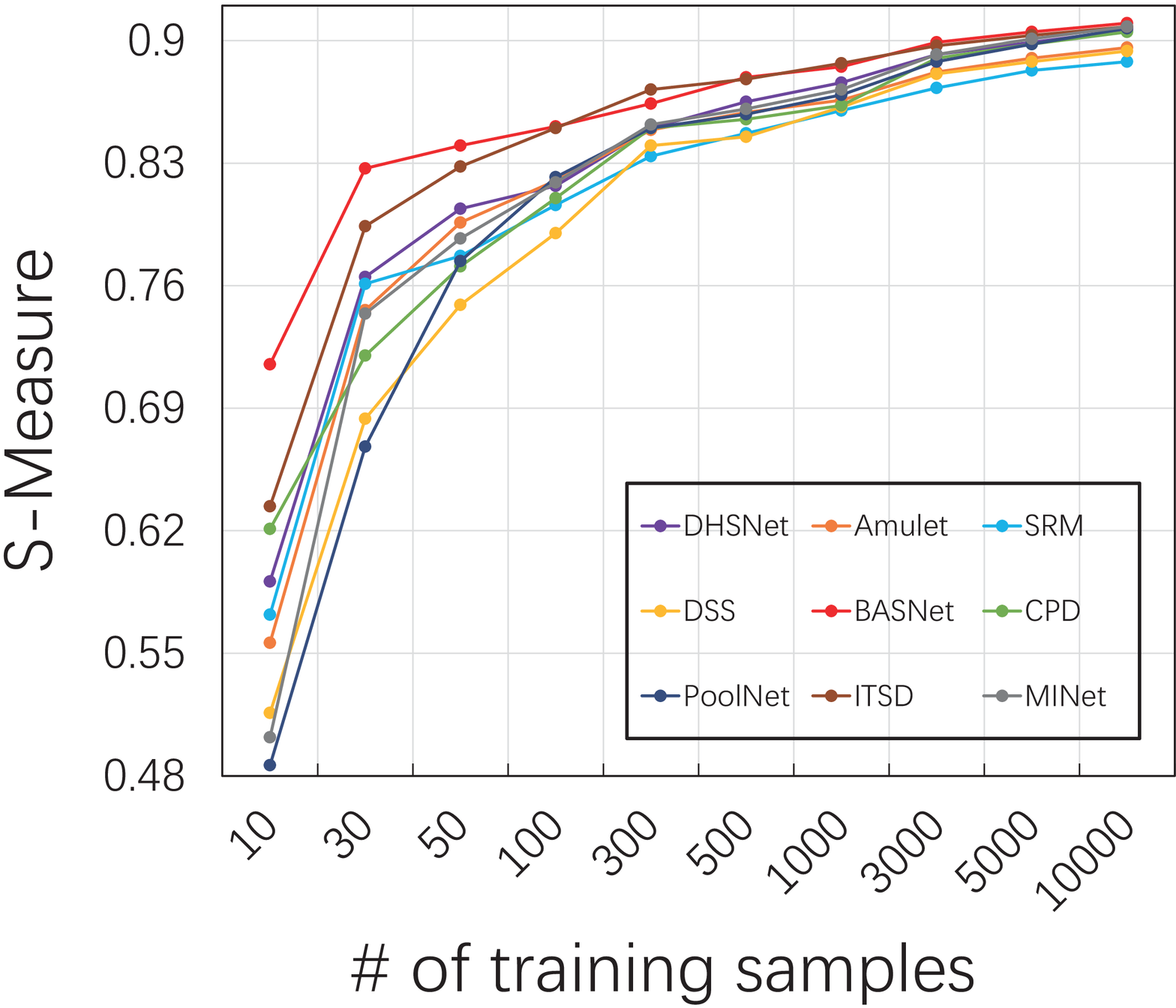}\hspace{20pt}
\includegraphics[width=0.46 \textwidth]{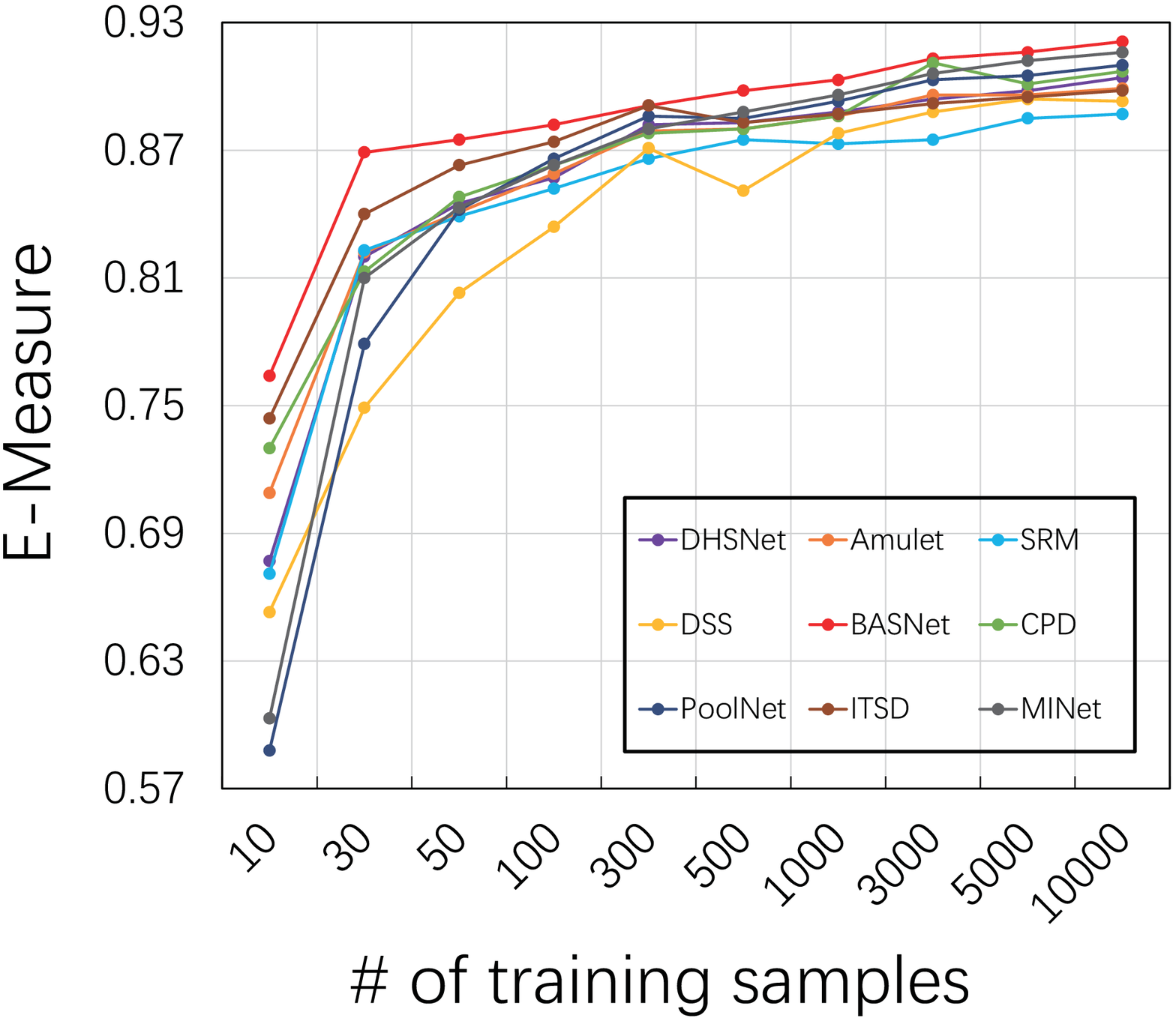}

\caption{ Curves of nine representative models when training on different numbers of images.}
\label{fig:fewshot}
\end{figure*}

\begin{table*}[!t]
\renewcommand\arraystretch{1.1}
\caption{Ablation study on loss functions using BASNet \cite{basnet} as baseline.
We compare our EA loss with other four losses used in previous methods.
``Others'' means the remaining 10 methods in our benchmark.
w$F_{\beta}$ indicates the proposed weighted $F_\beta$ loss.
Results of our loss are in bold.
}
\label{tab:loss}
\centering
\footnotesize
\begin{tabular}{ccccc|c|cccccc}
\toprule
BCE         & CTLoss     & SSIM       & IOU       & w$F_{\beta}$ & Used in &
max-$F_{\beta}$ & ave-$F_{\beta}$ & Fbw & MAE     & SM         & EM   \\\hline
\checkmark & --         & --         & \checkmark & --         & \cite{minet} \cite{nldf}      &
.917       & .861       & .880       & .034       & .909       & .920 \\
\checkmark & --         & \checkmark & \checkmark & --         & \cite{basnet}      &
.916       & .860       & .880       & .033       & .908       & .920 \\
--         & \checkmark & --         & --         & --         & \cite{itsd}       &
.916       & .831       & .848       & .040       & .910       & .905 \\
\checkmark & --         & --         & --         & --         & Others &
.915       & .838       & .856       & .038       & .910       & .909 \\\hline
--         & \checkmark & --         & --         & \checkmark & Ours &
\textbf{.918}& \textbf{.872} & \textbf{.880} & \textbf{.035} & \textbf{.909} & \textbf{.923} \\
\noalign{\smallskip}\bottomrule
\end{tabular}
\end{table*}

\begin{table*}[!t]
\caption{Quantitative results of our EA loss on the test set of the proposed SALOD benchmark. Settings for all models are the same as in the benchmarking experiment.}
\label{tab:gen}
\centering
\footnotesize
\begin{tabular}{lc|cccccc}
\toprule
Methods   & EA loss & max-$F_\beta$ & ave-$F_\beta$ & Fbw   & MAE   & SM   & EM        \\
\noalign{\smallskip}\midrule
\multirow{2}{*}{DHSNet\cite{dhsnet}} & \XSolidBrush & .909  & .830  & .848  & .039  & .905 & .905\\
                                     & \checkmark   & .911  & .860  & .869  & .036  & .905 & .918\\\hline
\multirow{2}{*}{Amulet\cite{amulet}} & \XSolidBrush & .895  & .822  & .835  & .042  & .894 & .900\\
                                     & \checkmark   & .898  & .847  & .852  & .040  & .893 & .911\\\hline
\multirow{2}{*}{SRM\cite{srm}}       & \XSolidBrush & .882  & .803  & .812  & .047  & .885 & .891\\
                                     & \checkmark   & .887  & .844  & .837  & .042  & .887 & .910\\\hline
\multirow{2}{*}{DSS\cite{dss}}       & \XSolidBrush & .894  & .821  & .826  & .045  & .893 & .898\\
                                     & \checkmark   & .900  & .859  & .853  & .039  & .897 & .919\\\hline
\multirow{2}{*}{BASNet\cite{basnet}} & \XSolidBrush & .917  & .861  & .884  & .032  & .909 & .921\\
                                     & \checkmark   & .918  & .868  & .882  & .034  & .911 & .922\\\hline
\multirow{2}{*}{CPD\cite{cpd}}       & \XSolidBrush & .906  & .842  & .836  & .040  & .904 & .908\\
                                     & \checkmark   & .906  & .869  & .856  & .039  & .902 & .919\\\hline
\multirow{2}{*}{PoolNet\cite{poolnet}}& \XSolidBrush & .912  & .843  & .861  & .036  & .907 & .912\\
                                     & \checkmark   & .913  & .867  & .875  & .035  & .906 & .922\\\hline
\multirow{2}{*}{ITSD\cite{itsd}}     & \XSolidBrush & .913  & .825  & .842  & .042  & .907 & .899\\
                                     & \checkmark   & .909  & .858  & .866  & .039  & .902 & .915\\\hline
\multirow{2}{*}{MINet\cite{minet}}   & \XSolidBrush & .913  & .851  & .871  & .034  & .906 & .917\\
                                     & \checkmark   & .916  & .868  & .880  & .034  & .909 & .923\\
\noalign{\smallskip}\bottomrule
\end{tabular}
\end{table*}

\subsection{Few-shot Experiment}
In this section, we analyze the impact of the number of training examples on nine representative SOD methods.
They are DHSNet \cite{dhsnet}, Amulet \cite{amulet}, SRM \cite{srm}, DSS \cite{dss}, BASNet \cite{basnet}, CPD \cite{cpd}, PoolNet \cite{poolnet}, ITSD \cite{itsd} and MINet \cite{minet}.
We select ten subsets with various scales to train these methods and evaluate their performance on the same test set with 27930 images.
The curves of six metrics on different train sets are visualized in Fig. \ref{fig:fewshot}.

Overall, experiment results prove that more training samples significantly improve the generalization ability of SOD methods.
Specifically, BASNet \cite{basnet} obtains the most stable results compared with other methods.
ITSD \cite{itsd} employs an interactive structure, which provids more supervision signals to promote the network to learn more generalized features.
Thus, it performs better than most of compared methods.
The results of BASNet and ITSD prove that contour information can help SOD methods to learn better representations from a few images.
As the number of training samples decreased, some methods report greater performance degradations, such as DSS \cite{dss}, CPD \cite{cpd} and Amulet \cite{amulet}.
Moreover, PoolNet \cite{poolnet} and MINet \cite{minet} report much lower scores using 10 samples, which prove that the sophisticated decoders are hard to conclude more generalized patterns from a few samples.
This observation indicates that these methods are more sensitive to train set scales.
The main reason is that their sophisticated decoders perform well with a bunch of training data, but they are prone to overfitting when only a limited number of training samples is available.

\section{Loss Function}
\subsection{Compared to Other Losses}
In this subsection, we select BASNet as the baseline due to its impressive performance in the above experiments.
To validate the effectiveness of loss functions, we employ several existing losses, such as BCE, CTLoss, SSIM, IOU and $F_{\beta}$, as well as the proposed EA loss to train BASNet.
We use some combinations of existing loss functions as implemented in existing SOD methods, including $L_{BCE}+L_{IOU}$ (MINet \cite{minet} and NLDF \cite{nldf}), $L_{BCE}+L_{IOU}+L_{SSIM}$ (BASNet \cite{basnet}), $L_c$ (ITSD \cite{itsd}), $L_{BCE}$ (others).
Moreover, we only supervise the final prediction of these networks to reduce the variance caused by intermediate supervision.
Experiment results are shown in Tab. \ref{tab:loss}.
Our EA loss achieves the best max-$F_\beta$, ave-$F_\beta$, Fbw and E-measure scores, and also obtains very competitive MAE and S-Measure scores against others.

\subsection{Comparisons using More Models}
To prove the generalization ability of the proposed EA loss, we list the results of nine representative SOD methods in Tab. \ref{tab:gen}.
We draw four conclusions based on this table.
First, for DHSNet \cite{dhsnet}, SRM \cite{srm}, DSS \cite{dss} and MINet \cite{minet}, the proposed EA loss reports better results compared with their original loss functions.
Second, for Amulet \cite{amulet}, CPD \cite{cpd} and PoolNet \cite{poolnet}, we can observe large improvements on most metrics.
Third, ACT loss in ITSD \cite{itsd} surpasses our loss on the max-$F_{\beta}$ and S-Measure scores because it makes the network pay more attention on boundary pixels.
Last, we observe slightly better performance on BASNet \cite{basnet} using our loss.
Overall, these experiments demonstrate that the proposed EA loss achieves a more robust performance compared with existing losses.

\section{Conclusion}
In this work, we propose a novel benchmark to impartially evaluate the performance of 14 representative SOD methods by eliminating their implementation discrepancy.
Specifically, we collect a hybrid dataset based on several prevalent SOD datasets and employ multiple effectiveness and efficiency metrics to comprehensively evaluate these SOD methods.
Moreover, two additional protocols are set up to validate the performance of these methods in some limited conditions, such as very few training samples or large objectness shifting between train and test sets.
Some conclusions can be draw in our benchmark.
First, some earlier methods benefit from the consistent settings and thus obtain significantly improvements.
Second, objectness shifting between train and test sets brings a great challenge to all available SOD methods in our benchmark.
Last, these methods also fail to learn more discriminative representations with only a few samples.
We expect that our training and testing protocols provide the community with a reliable and consistent evaluation.

Based on the above experiments, we proposed a novel Edge-Aware (EA) loss to promote the networks to conclude more distinctive representations.
Experiments prove that our EA loss can assist existing SOD methods to achieve better performance than their original losses.
This seems to be a promising future direction to improve the performance of SOD methods.

\section*{Acknowledge}
This work was supported in part by the Key-Area Research and Development Program of Guangdong Province under Grant 2019B010155003, the National Natural Science Foundation of China under Grant 62072482, and the Guangdong-Hong Kong-Macao Greater Bay Area International Science and Technology Innovation Cooperation Project under Grant 2021A0505030080.



\bibliographystyle{sn-mathphys}
\bibliography{sn-bibliography}


\end{document}